\documentclass[acmsmall]{acmart}
\AtBeginDocument{%
  \providecommand\BibTeX{{%
    \normalfont B\kern-0.5em{\scshape i\kern-0.25em b}\kern-0.8em\TeX}}}

\setcopyright{acmcopyright}
\copyrightyear{2022}
\acmYear{2022}
\acmDOI{XXXXXXX.XXXXXXX}

\acmJournal{TODAES}
\acmVolume{37}
\acmNumber{4}
\acmArticle{111}
\acmMonth{8}

\usepackage{soul}
\usepackage{xcolor}
\usepackage{xspace}
\usepackage{booktabs}
\usepackage[flushleft]{threeparttable} 
\usepackage{bm}
\usepackage{wrapfig}
\usepackage{multirow}
\usepackage{multicol}
\usepackage{amsmath}
\usepackage{amsfonts}
\usepackage{enumitem}
\usepackage{color}
\usepackage{xcolor}
\usepackage{soul}
\usepackage{extarrows}
\usepackage{gensymb}
\usepackage{natbib}
\usepackage{microtype}
\usepackage{graphicx}
\usepackage{subcaption}
\usepackage{stfloats} 
\usepackage[bottom]{footmisc}
\newcommand*\circled[1]{\tikz[baseline=(char.base)]{
            \node[shape=circle,fill,inner sep=1.2pt] (char) {\textcolor{white}{#1}};}}
\newcommand{\tablefontsize}[0]{
\fontsize{8pt}{10pt}\selectfont
}
\usepackage[ruled,linesnumbered]{algorithm2e}
\newlength\mylenin
\newcommand\myinput[1]{%
\settowidth\mylenin{\KwIn{}}%
\setlength\hangindent{\mylenin}%
\hspace*{\mylenin}#1\\}

\let\oldnl\nl
\newcommand{\nonl}{\renewcommand{\nl}{\let\nl\oldnl}}

\newlength\mylenout

\usepackage{algpseudocode}
\DeclareMathOperator*{\argmin}{argmin}
\usepackage{tikz}

\newif\ifcomment
\commenttrue
\commentfalse 
            
\ifcomment
\definecolor{stelios_colour}{RGB}{191, 232, 255}
\newcommand{\stelios}[1]{\sethlcolor{stelios_colour}\hl{[Stelios: #1]}}

\definecolor{javier_colour}{RGB}{255, 204, 204}
\newcommand{\javier}[1]{\sethlcolor{javier_colour}\hl{[Javier: #1]}}

\else
\newcommand{\stelios}[1]{}
\newcommand{\javier}[1]{}

\fi

\newcommand{\rev}[1]{\textcolor{black}{#1}}

\newcommand{\tool}{unzipFPGA\xspace}

\begin{document}

\title{Mitigating Memory Wall Effects in CNN Engines with On-the-Fly Weights Generation}


\author{Stylianos I. Venieris}
\authornote{Both authors contributed equally to this research.}
\email{s.venieris@samsung.com}
\orcid{1234-5678-9012}
\affiliation{%
  \institution{Samsung AI Center, Cambridge}
  \country{UK}
}
\author{Javier Fernandez-Marques}
\authornotemark[1]
\email{j1.fernandez@samsung.com}
\orcid{1234-5678-9012}
\affiliation{%
  \institution{Samsung AI Center, Cambridge}
  \country{UK}
}
\author{Nicholas D. Lane}
\orcid{1234-5678-9012}
\affiliation{%
  \institution{Samsung AI Center, Cambridge \& University of Cambridge}
  \country{UK}
}
\email{nic.lane@samsung.com}

\renewcommand{\shortauthors}{Venieris and Fernandez-Marques, et al.}

\ccsdesc[500]{Computer systems organization~Reconfigurable computing}
\ccsdesc[500]{Computing methodologies~Neural networks}

\keywords{neural networks, hardware accelerator, weights generation}


\begin{abstract}
    The unprecedented accuracy of convolutional neural networks (CNNs) across a broad range of AI tasks has led to their widespread deployment in mobile and embedded settings. In a pursuit for high-performance and energy-efficient inference, significant research effort has been invested in the design of FPGA-based CNN accelerators. In this context, single computation engines constitute a popular design approach that enables the deployment of diverse models without the overhead of fabric reconfiguration. Nevertheless, this flexibility often comes with significantly degraded performance on memory-bound layers and resource underutilisation due to the suboptimal mapping of certain layers on the engine's fixed configuration. 
In this work, we investigate the implications in terms of CNN engine design for a class of models that introduce a pre-convolution stage to decompress the weights at run time. We refer to these approaches as \textit{on-the-fly}. 
This paper presents \tool, a novel CNN inference system that counteracts the limitations of existing CNN engines. The proposed framework comprises a novel CNN hardware architecture that introduces a weights generator module that enables the on-chip on-the-fly generation of weights, alleviating the negative impact of limited bandwidth on memory-bound layers. We further enhance \tool with an automated hardware-aware methodology that tailors the weights generation mechanism to the target CNN-device pair, leading to an improved accuracy-performance balance.
Finally, we introduce an input selective processing element (PE) design that balances the load between PEs in suboptimally mapped layers. Quantitative evaluation shows that the proposed framework yields hardware designs that achieve an average of 2.57$\times$ performance efficiency gain over highly optimised GPU designs for the same power constraints and up to 3.94$\times$ higher performance density over a diverse range of state-of-the-art FPGA-based CNN accelerators.

\end{abstract}

\maketitle

\section{Introduction}
\label{sec:intro}

The unparalleled accuracy of convolutional neural networks (CNNs) across a broad range of AI inference tasks has led to the development of novel applications. With a large portion deployed across mobile and embedded devices, there is a need for high-performance, energy-efficient implementations that can deliver responsiveness and prolonged battery life. Currently, the conventional computing platforms for CNN inference comprise either CPUs and GPUs~\cite{fb_edge2019hpca,oodin2021smartcomp,embench2019emdl} or custom application-specific integrated circuits (ASICs), such as neural processing units (NPUs)~\cite{embench2019emdl, ai_benchmark2019iccvw,samsung_npu2021isca}. On the one hand, CPU- and GPU-based systems can support diverse CNN models through their programmability, but penalise performance in order to provide this generality~\cite{embench2019emdl}. On the other hand, ASICs provide significant acceleration under a minimal power envelope~\cite{ai_benchmark2019iccvw}. Nevertheless, the benefits of ASICs typically require the functionality to remain fixed after fabrication, leaving no room for applying model-specific optimisations~\cite{residaccel2017iscas,mobilenet_accel2021fpl} or mapping newer CNN models~\cite{fpgaconvnet2019tnnls}.

To provide a balance between flexibility and high performance, numerous CNN accelerators target reconfigurable hardware platforms, such as field-programmable gate arrays (FPGAs). Currently, the FPGA-based accelerator landscape spans a wide spectrum, from flexible CNN-specific processors~\cite{snowflake2017iscas,lightopu2020fpga,uniopu2020tvlsi} to highly customised streaming architectures~\cite{streaming2016fpl,fpgaconvnet2019tnnls,finn2017fpga,lutnet2020toc}. One of the most widely adopted paradigms that lies in the midpoint of the flexibility-customisation spectrum is the single computation engine (SCE)~\cite{fpdnn2017fccm,dla2018fpl,cascadecnn2020date,caffeine2019tcad,dnnvm2019tcad,abdelfattah2020best,gamma2020iccad,cascadecnn2018fpl,fft2020fpga,alamo2020tcad}. Under this paradigm, a powerful processing engine is time-shared to sequentially execute the layers of a CNN. 
This allows for accelerator's resources to be reused across both layers and CNN models, without the need to reconfigure the fabric.

Despite the flexibility of SCEs, their attainable performance is often bounded by two primary factors:
\textit{1)}~layers with low computation-to-communication ratio that become memory-bound~\cite{cascadecnn2018fpl,fft2020fpga,alamo2020tcad} and \textit{2)}~the suboptimal mapping of diverse layers on the fixed configuration of the SCE that leads to underutilised processing elements (PEs)~\cite{alamo2020tcad,latency2017fpl,cnnroofline2015fpga}. These two obstacles set a hard limit to the actual sustained performance that this family of accelerators can achieve, indicating an emerging need for novel solutions to counteract their impact.

Concurrently with the continuing hardware advances, a growing body of work focuses on compressing CNNs through a class of lossy non-structural methods~\cite{ha2016hypernetworks, pmlr-v80-qiu18a, fernandez2018binarycmd, ijcai2018-380, ovsf2018emdl, Yang2020FSNet}.
Orthogonally to other model simplification techniques such as pruning or quantisation, this group of methods dictates that the weights of a model are deployed in a compact form and are ``inflated" only at run time. Given that several CNN layers are constrained by the limited off-chip memory bandwidth of the target computing platforms~\cite{latency2017fpl,cascadecnn2018fpl,fft2020fpga,alamo2020tcad}, storing the compressed weights on-chip and reconstructing them on-the-fly can play a decisive role in alleviating the memory-boundedness and enabling the better utilisation of the computational resources.

Nevertheless, the novel dataflow and execution scheme 
of such models brings up a new challenge regarding their optimised mapping.
Existing accelerators have been designed for conventional deep models, adopting either a streaming or layer-by-layer execution~\cite{cnnfpgatoolflows2018csur}.
Hence, despite the significant potential of on-the-fly models, their different execution paradigm renders conventional architectures futile in serving them. 

This paper presents a novel CNN system that overcomes the withstanding limitations of single computation engines and enables the efficient and high-performance execution of on-the-fly models. At the core of the proposed system lies a novel CNN hardware architecture. To alleviate the impact of the memory wall, we introduce a hardware-based weights generator that is responsible for efficiently generating the CNN weights on-the-fly. 
Comprising a custom memory organisation and a highly optimised datapath, the weights generator is scalable, with tunable parameters that allow it to be tailored to the needs of the target application, the workload characteristics of the CNN and, the capabilities of the FPGA device.
To counteract the underutilisation of computational resources due to the suboptimal mapping of diverse layers, we further propose a novel CNN engine design comprising input selective PEs. Under this design, a subset of PEs is enhanced with efficient switches that enable neighbouring PEs to perform load-balancing through seamless work-stealing.
Finally, we present a framework for deriving on-the-fly models from pre-trained CNNs and mapping them on a given FPGA.

The initial work in~\cite{unzipfpga2021fccm} provided a high-level overview of the proposed framework and its on-the-fly weights formulation.
Moreover, the evaluation solely focused on end-to-end performance compared to other FPGA works.
In this paper, we first present the complete framework, encompassing both the algorithmic and hardware aspects (Section~\ref{sec:design_flow}). As such, we provide a detailed description of the on-the-fly weights generation process and its algorithmic underpinning on OVSF codes (Section~\ref{sec:background_cnn_ovsf}), and we position the proposed hardware architecture with respect to the status-quo CNN engines (Section~\ref{sec:arch}).

Moreover, in the initial work, the compression ratios of each CNN layer were manually selected based on heuristics. In this work, we make steps towards automation by introducing a hardware-aware scheme for tuning the per-layer compression ratios of OVSF models (Section~\ref{sec:hw_aware_ratios}). The proposed method exploits the bottleneck characteristics of each layer in order to generate more accurately its weights while sustaining high throughput, leading to a better accuracy-performance trade-off (Section~\ref{sec:eval_hw_aware_ratios}).

Finally, we present for the first time an in-depth evaluation of various critical aspects of the proposed framework. These include the evaluation of two techniques to derive on-the-fly models from vanilla CNNs (Section~\ref{sec:basis_selection_and_3x3_extraction}), a thorough study of the impact of the proposed input selective PEs (Section~\ref{sec:input_sel_pes_eval}), an extensive comparison with highly optimised designs on embedded GPUs (Section~\ref{sec:gpu}) and further comparisons with the most recent state-of-the-art FPGA-based accelerators (Section~\ref{sec:fpga_comparison}).

\section{Background and Related Work}
\label{sec:background}

This section first positions the proposed on-the-fly formulation in context with existing optimisation techniques for CNNs. We continue with a brief historical context for OVSF codes and how to construct them. Then, we show how orthogonal variable spreading factor (OVSF) codes are used to construct filters in CNNs and, how they are integrated into the training process. This section concludes with an overview of the challenges and opportunities of CNN-targeting single computation engines, which is our paradigm of choice when implementing on-the-fly models on FPGAs.

\subsection{Designing Lightweight Convolutional Neural Networks}

The plethora of existing techniques to modify CNNs for faster inference can be categorised into: pruning~\cite{luo2017thinet, blalock2020state, he2018soft}, which removes redundant parameters; 
quantisation~\cite{krishnamoorthi2018quantizing,jacob2018quantization,dong2019hawq,fernandezmarques2020searching,alizadeh2018a}, which results in compact low-precision models; or, sparsification~\cite{wen2016learning, sparsify2016sensys, gale2020sparse}, which leverages compressed data formats. In addition, a number of frameworks combine several of these techniques. Most notably: Deep Compression~\cite{deepcompression2015iclr} which, given an over-parametrised model, 
applies pruning, quantisation and Huffman encoding;  RedCNN~\cite{pmlr-v70-wang17m} which prunes channels based on an activation overlap metric; 
and, more recently, APQ~\cite{wang2020apq}, which designs a CNN that meets given computational, memory and latency constraints through a joint optimisation formulation. 

\subsection{On-the-Fly Convolutional Neural Networks}
\label{sec:background_cnn_ovsf}
Orthogonal to these methods, various works have explored ways of obtaining extremely compact model representations by factorising the filters in CNNs or by passing these through a multi-stage compression pipeline. Common to these methods is the need for a \textit{decompressing} stage that generates the filters at run time during inference. 
A selection of such 
techniques include: \cite{ha2016hypernetworks} that uses an auxiliary NN to generate each layer's weights in the main network given an embedding of the weights.  
In~\cite{pmlr-v80-qiu18a}, weight filters are constructed as a dense combination of a set of Fourier Bessels bases that are generated deterministically at run time.
Another technique exploiting deterministic bases was presented in~\cite{fernandez2018binarycmd,ijcai2018-380, ovsf2018emdl}, where bases are formed from OVSF binary codes. 
It enables the construction of model weights by learning a linear combination of OVSF bases during training. 

We label these techniques as \emph{on-the-fly} since they \textit{1)} require a single-step decompression stage to obtain the filters, \textit{2)} this process can be done on-demand, \textit{i.e.}~at each layer and for every input, and \textit{3)} such decompression is lightweight, \textit{i.e.}~it does not require multiple inferences to amortise its associated costs. In this work, we focus on on-the-fly methods that use compression as a means to reduce data-movement overheads, balancing the costs of decompressing the model parameters with the latency savings due to reduced off-chip or main memory accesses.

To this end, this work makes use of OVSF codes to compress filters in a CNN and reconstruct them during inference in a lightweight manner. Three additional reasons motivated the choice of this class of codes: 
\textit{1)}~OVSF codes are \textit{binary} and thus can be efficiently stored on-chip~\cite{5291322}; \textit{2)}~their theoretical properties are well studied by the wireless community~\cite{Andreev:2003:OCG:764808.764868,1411169}; \textit{3)}~they offer good compression-accuracy trade-off (i.e. lossy compression) in various AI tasks~\cite{ovsf2018emdl,ijcai2018-380}. 
Nonetheless, \cite{ovcf-fpga}~is the only existing FPGA-based OVSF design, presenting solely a direct implementation specific for communication systems. To effectively use OVSF with CNNs, the underlying hardware design needs to be tailored and optimised for the CNN dataflow.

\subsection{On-the-Fly OVSF CNN Layers}
The chosen OVSF codes are a set of mutually orthogonal binary codes, originally designed to split in the frequency domain signals from different users in W-CDMA-based 3G cellular systems~\cite{wdscdma}. Using them as channelisation codes allowed for communication channels to remain orthogonal in multi-user access scenarios, reducing signal interference while dramatically increasing the system capacity.

These codes can be obtained using Sylvester's construction algorithm for Hadamard matrices. In this way, given $H_0 = [1]$ and $H_2$, subsequent $H_{2^k}$ expansions are defined as
%
\begin{equation}
  {H_{2}} =
  \begin{bmatrix}
    1 & 1 \\
    1 & -1
  \end{bmatrix}
  , \quad
  {H_{2^k}} =
  \begin{bmatrix}
    H_{2^{k-1}} & H_{2^{k-1}} \\
    H_{2^{k-1}} & -H_{2^{k-1}}
  \end{bmatrix}
  =
  {H_2 \otimes H_{2^{k-1}}}
\label{eq:sylvester}
\end{equation}
where $H_{2^k}$ is an $L$$\times$$L$ Hadamard matrix, with $L = 2^k, k\in \mathbb{N}$ and $\otimes$ is the Kronecker product. For $k>1$, each row is an OVSF code fulfilling the properties of being binary and orthogonal to each other. This will enable us to use them as basis for $\mathbb{R}^L$. An alternative formulation~\cite{1997iet} allows for the construction of OVSF codes as a recursive expansion of a perfect binary tree.

By treating the set of $L$ OVSF codes as a basis spanning $\mathbb{R}^L$, we can define the construction process of an arbitrary real-valued vector $\pmb{v}'_i$ as the linear combination of such codes
\begin{figure*}[t]
    \centering
    {
    \includegraphics[width=0.8\linewidth]{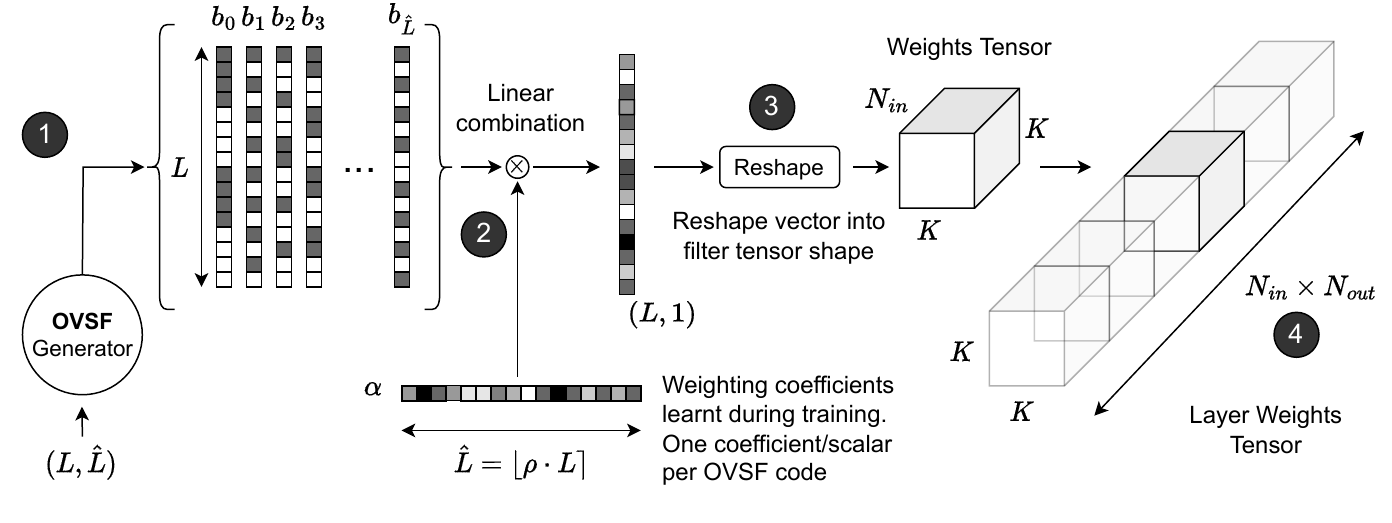}
    }
    \vspace{-0.1cm}
    \captionsetup{font=small,labelfont=bf}
    \caption{\footnotesize Constructing filters of a convolutional layer using OVSF codes. With an OVSF code generator outputing binary codes of length $L$ $\bigl(\protect\circled{1}\bigr)$, a $K$$\times$$K$ filter with $N_{\text{in}}$ channels (\textit{i.e.}~of shape $N_{\text{in}}$$\times$$K$$\times$$K$) is obtained by performing a linear combination $\bigl(\protect\circled{2}\bigr)$ of $\hat{L} = \lfloor \rho \cdot L\rceil$ codes of length $L=N_{\text{in}}$$K$$K$, with $\rho \in [0,1]$. Then, the resulting $L$$\times$$1$ vector is reshaped to match the target filter's shape $\bigl(\protect\circled{3}\bigr)$. For a convolutional layer with $N_{\text{out}}$ output channels, this process is repeated $N_{\text{out}}$ times, concatenating the results $\bigl(\protect\circled{4}\bigr)$.}
    \label{fig:ovsf_gen_diagram}
\end{figure*}
\begin{equation}
  \pmb{v}'_i =  \sum_{j = 0}^{\lfloor \rho \cdot L \rceil}\alpha^{j}_{i}\bm{b}^{j}_{i}
  ,
  \quad
  E_{i} = \left\|\bm{v}'_i - \bm{v}_i \right \|_2^2 = \left \| \sum_{j = 0}^{\lfloor \rho \cdot L \rceil}\alpha^{j}_{i}\bm{b}^{j}_{i} - \bm{v}_i \right \|^2_2 < \epsilon
\label{eq:vector_eq}
\end{equation}
\vspace{1mm}
where $\pmb{\alpha}_i = \{\alpha^{0}_i, \alpha^{1}_i, \alpha^{2}_i, ... , \alpha^{L-1}_i\}$ are weighting coefficients, $\pmb{b}^{j}_i$ is the $j$-th OVSF binary code of length $L$ and, $\rho \in [0,1]$ is the ratio of codes to use in order to construct $\pmb{v}_i$. The expression on the right measures the difference between $\pmb{v}'_i$ and a real-valued standard vector of length $L$, $\pmb{v}_i$. Intuitively, $\mathlarger{\epsilon} \to 0$ as we increase the ratio of binary codes used. 

When constructing matrices from OVSF codes or higher-dimensional tensors, a reshaping stage follows the linear combination shown in Eq.~(\ref{eq:vector_eq}). In this way, if the weights tensor of a given convolutional layer is of shape $N_{\text{out}}$$\times$$N_{\text{in}}$$\times$$K$$\times$$K$, the construction process using OVSF could be framed as the concatenation of $N_{\text{out}}$ $N_{\text{\text{in}}}$$\times$$K$$\times$$K$ filters using codes of length $L=N_{\text{in}}$$K$$K$ and up to $L$ of such codes. Here, $N_{\text{in}}$ and $N_{\text{out}}$ stands for the number of input and output channels in the layer respectively, and $K$$\times$$K$ are the spatial dimensions of the convolutional filter. This scenario is illustrated in Fig.~\ref{fig:ovsf_gen_diagram}. The set of scalars $\{\alpha \}_{k=1}^L$ in each layer are learnt via standard backpropagation and represent the only learnable parameters in an OVSF layer since the OVSF binary codes themselves are fixed. A compressed representation of $f_{i}$ is obtained when $\rho$$<$$1$.
Upon deployment, the filters are first generated and then the main inference computation proceeds as normal.

Due to the construction process of OVSF codes (Eq.~\ref{eq:sylvester}), $K$ in OVSF convolutional layers is limited to be a power-of-two number. We address this in Sec.~\ref{sec:OVSF_issues}, where we present two methods to enable the construction of filters with $K\in\mathcal{N}$, for example with the ubiquitous $K=3$.

\subsection{Challenges of FPGA-based CNN Engines}

Until now, a wide array of FPGA-based CNN accelerators have been proposed. In the customisation-programmability spectrum, existing designs span from custom streaming architectures~\cite{streaming2016fpl,fpgaconvnet2019tnnls} and accelerators for quantised~\cite{finn2017fpga,ternaryfpga2018trets,finnr2018trets,multiprec2019fpt,lutnet2020toc,logicnets2020fpl} and sparse CNNs~\cite{cambriconx2016micro,cambircons2018micro,scnn2017isca,sparten2019micro,sparsecnnaccel2019fccm,extensor2019micro,tensaurus2020hpca}, up to instruction-based processors~\cite{snowflake2017iscas,lightopu2020fpga,uniopu2020tvlsi}.
One of the most well adopted paradigms are the single computation engines~\cite{fpdnn2017fccm,dla2018fpl,cascadecnn2020date,caffeine2019tcad,dnnvm2019tcad,cascadecnn2018fpl,fft2020fpga,alamo2020tcad,abdelfattah2020best}, due to their balanced trade-off of programmability and performance. 
Currently, despite the progress in processing unit design, further gain in the attainable performance of such engines is hindered by two main factors: i)~memory-bound layers that are dominated by the communication with the external memory~\cite{mem_req2918iiswc,cascadecnn2018fpl,fft2020fpga,alamo2020tcad}. While embedded platforms provide limited bandwidth~\cite{lostbw2019fpt,divrsemem2019fpt,venieris2018fpl}, \textit{e.g.} less than 4.5~GB/s for Ultra96 and ZC706, sustaining peak bandwidth even on larger devices, such as ZCU104, is nontrivial~\cite{divrsemem2019fpt}. This is aggravated as multiple applications are collocated on a single device~\cite{venieris2018fpl,codesignmultiple2020dac,multinn2020isca}; and ii)~underutilised PEs due to the mismatch of diverse layer shapes~\cite{alamo2020tcad,latency2017fpl,cnnroofline2015fpga,gamma2020iccad,abdelfattah2020best,maestro2020micro}. 

\textbf{Memory-centric Designs.}
The memory bandwidth problem faced by CNN engines has been studied in previous work. 
EIE~\cite{eie2016isca} uses the Deep Compression method~\cite{deepcompression2015iclr} to compress the weights of fully connected (FC) layers. However, as FC layers have been mostly abandoned in modern CNNs, its applicability is limited.
Angel-Eye~\cite{angeleye2018tcad} compresses all layers through precision quantisation. 
Cambricon-X~\cite{cambriconx2016micro} transfers only the non-zero weights, while Cambricon-S~\cite{cambircons2018micro} and Scalpel~\cite{scalpel2017isca} apply coarse weight pruning, but with significant accuracy drop.
CircCNN~\cite{circcnn2017micro} uses block-circulant matrices for weights, but requires complex FFT hardware for efficient execution.
\cite{permdnn2018micro} converts sparse weights to permuted diagonal matrices, but only targets FC layers. 
\cite{escher2017fccm}~exploits large batch sizes to increase weights reuse and, thus, is not suitable for latency-critical applications that cannot tolerate batching~\cite{latency2017fpl}.

Focusing on activations, \cite{fusedlayer2016micro} fuses adjacent layers to cache intermediate activations, while \cite{scnn2017isca} and Eyeriss~\cite{eyeriss2017jssc}, \cite{def2021aspdac} employ encoding schemes to minimise their bandwidth footprint.
Other solutions have either relied on large devices~\cite{opencldla2017fpga} and multiple FPGAs~\cite{brainwave2018isca} to fit all weights on-chip, or utilised highly customised designs to exploit multi-precision cascades~\cite{cascadecnn2020date} or fine-grained pruning~\cite{sparsecnnaccel2019fccm} at the cost of notable accuracy drop.

\textbf{Tackling PE Underutilisation.}
So far, a limited number of designs have focused on ii). \cite{maximising2017isca} addresses underutilisation by grouping CONV layers based on the compatibility of their shapes. 
\cite{streaming2016fpl} maps each layer to a dedicated compute stage, which can be used only for shallower networks, but does not scale to the deeper models of today. Furthermore, a limited number of works rely on FFT-based designs with flexible dataflow~\cite{fft2020fpga} and costly ASIC solutions~\cite{flexflow2017hpca,recpatterns2017tvlsi,maeri2018asplos,polymorph2021tc} with highly flexible PE interconnect.

In contrast to these works, we propose an approach that is independent of the CNN engine by not requiring any modification to the engine architecture itself. \tool can benefit any existing single computation engine by augmenting it with its hardware weights generator and enhancing its PE array with lightweight switches, without affecting the PE's internal processing units. As such, \tool is orthogonal and complementary to quantisation~\cite{angeleye2018tcad}, activations' encoding~\cite{eyeriss2017jssc,scnn2017isca}, fusion~\cite{fusedlayer2016micro} and zero-skipping PEs~\cite{cnvlutin2016isca,pragmatic2017micro,cambircons2018micro,shapeshifter2019micro,sparten2019micro}.

\begin{wrapfigure}{R}{0.5\textwidth}
    \vspace{-0.4cm}
    \centering
    {
    \includegraphics[trim={0.8cm 6.5cm 17.25cm 1.5cm},clip,width=0.5\textwidth]{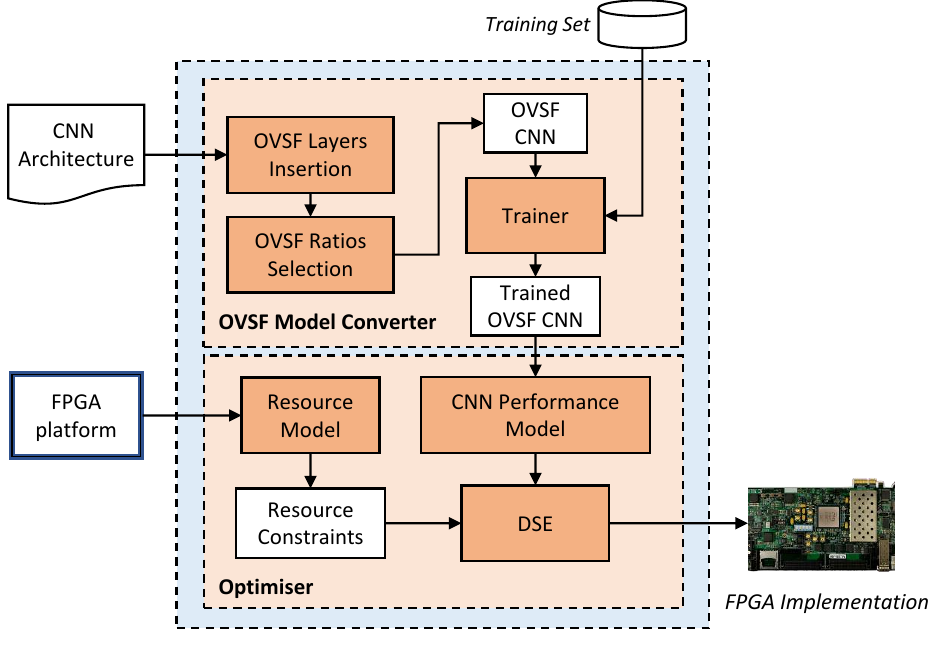}
    \vspace{-0.6cm}
    }
    \captionsetup{font=small,labelfont=bf}
    \caption{\footnotesize Overview of \tool's design flow.}
    \label{fig:design_flow}
     \vspace{-0.2cm}
\end{wrapfigure}

\section{\lowercase{unzip}FPGA's Design Flow}
\label{sec:design_flow}

Our framework aims to enhance the performance of hardware CNN engines, while maintaining a high level of abstraction for deep learning developers. Fig.~\ref{fig:design_flow} shows a high-level view of \tool's design flow, comprising two software components: \textit{1)}~the \textit{OVSF Model Converter} and \textit{2)}~the \textit{Optimiser}.

As a starting point, the deep learning expert provides the CNN model, expressed in PyTorch, and the target FPGA platform. The \textit{Converter} processes the supplied CNN architecture and derives an OVSF variant, by transforming the conventional convolutional layers into OVSF convolutional (OVSF-CONV) layers. This step entails \textit{i)}~the replacement of weight filters with a trainable linear combination of OVSF bases, followed by \textit{ii)}~the selection of each layer's compression ratio $\rho$. Next, the OVSF model is passed to the \textit{Trainer}, where the model gets trained using the supplied training set.

The \textit{Optimiser} accepts the trained OVSF CNN and a given FPGA platform and, uses them to populate the CNN \textit{Performance Model} and the \textit{Resource Constraints}, respectively. Importantly, the \textit{Optimiser} navigates the hardware configuration space considering resource allocations between the CNN engine and the weights generator. Upon completion, the design space exploration (\textit{DSE}) stage yields the highest performing configuration of \tool's architecture for the given CNN-device pair and the system is deployed on the FPGA.

\section{CNN Engine Design for On-the-Fly Weights}
\label{sec:arch}

This section starts by reviewing the hardware architecture of a conventional CNN engine, similar to the ones in \cite{shidiannao2915isca,eyeriss2017jssc,cascadecnn2018fpl}. Then, it provides a series of design requirements to enable on-the-fly weights generation for CNNs and presents our techniques for achieving them.

\subsection{Conventional CNN Engine Design}
\begin{wrapfigure}{R}{0.5\textwidth}
    \centering
    {
    \includegraphics[width=0.5\textwidth]{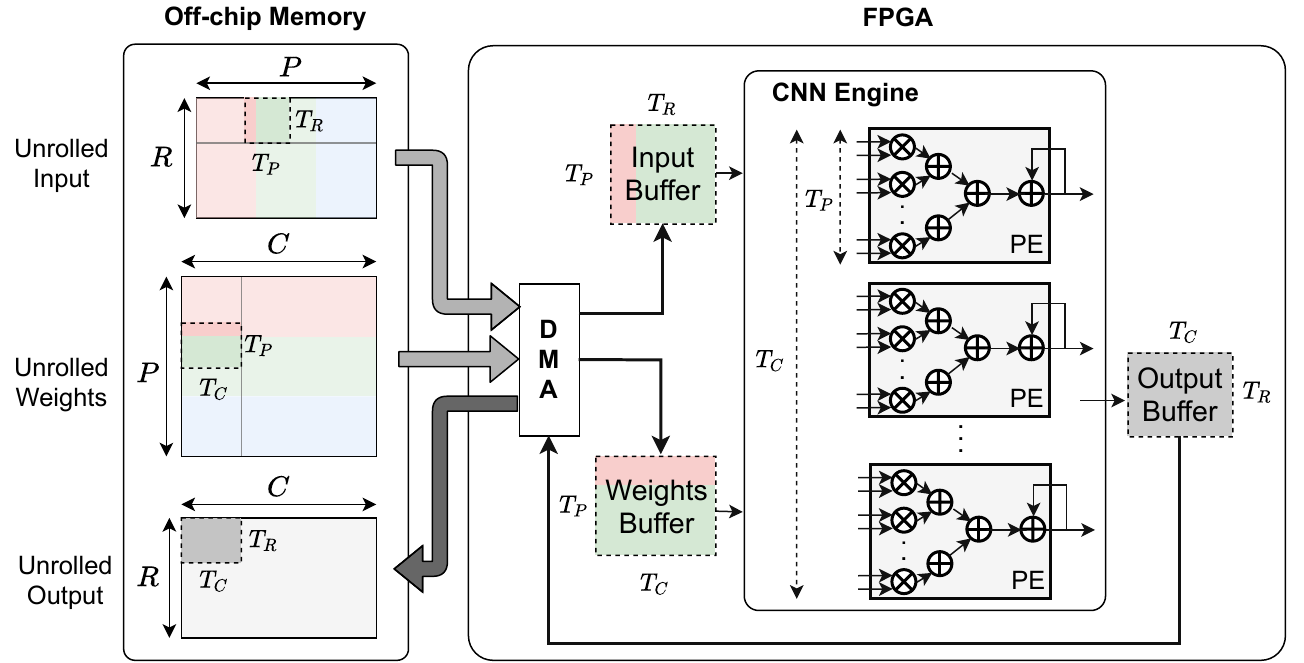}
    \vspace{-0.6cm}
    }
    \captionsetup{font=small,labelfont=bf}
    \caption{A conventional CNN engine.}
    \label{fig:conventional_cnn_engine}
     \vspace{-0.2cm}
\end{wrapfigure}

Fig.~\ref{fig:conventional_cnn_engine} illustrates a typical CNN engine design. The accelerator consists of an array of processing elements (PEs) to perform matrix multiplications and convolutions, one input and one output activations buffer, and a weights buffer. From an operational perspective, the CNN layers are scheduled sequentially, with pipelining applied between I/O communication and computation to hide the off-chip memory transfer latency.

\textbf{Processing Engine:}
To execute layers of various shapes and types, the core processing engine comprises an array of PEs for the execution of block matrix multiply (GEMM). Each PE contains a scalable dot-product circuit with configurable number of multiply-accumulate (MAC) units.
By translating convolutions into matrix multiplication, the engine can process both CONV and FC layers. To this end, a CONV layer with $N_{\text{in}}$ $H$$\times$$W$ input activations, $N_{\text{out}}$ output channels, $K$$\times$$K$ filters, $p$ padding and $S$ stride involves the multiplication between an $R$$\times$$P$ activations matrix and a $P$$\times$$C$ weights matrix to produce an $R$$\times$$C$ output matrix, with $R$$=$$\left\lceil \frac{H + 2p - K}{S}+1 \right\rceil \left\lceil \frac{W + 2p - K}{S} + 1  \right\rceil$, $P$$=$$N_{\text{in}} K^2 $ and $C$$=$$N_{\text{out}}$.

\textbf{Design-time Parametrisation:}
The CNN engine can be scaled based on the workload characteristics and the resources of the target FPGA. As such, it is parametrised with respect to the parameter tuple $\left<T_R,T_P,T_C\right>$. Each parameter determines the tile sizes for each matrix dimension $\left<R,P,C\right>$, the number of PEs ($T_C$) and the MAC units within each PE ($T_P$).

\textbf{Operational Flow:}
To produce a {\small $T_R$$\times$$T_C$} output tile, {\small $\left\lceil \frac{P}{T_P}\right\rceil$} tiles from the activations and weights matrices are processed and accumulated sequentially. 
A common mapping strategy (Fig.~\ref{fig:conventional_cnn_engine}) ties $T_P$ to the MACs per PE to exploit the parallelism within each $T_P$-wide dot product, and $T_C$ to PEs to parallelise the dot products at each output column. Overall, the rows of the {\small $T_R$$\times$$T_P$} activations tile are processed in a pipelined manner to maximise throughput. This is equivalent to an output stationary dataflow~\cite{shidiannao2915isca,eyeriss2017jssc,cascadecnn2018fpl}, which minimises the memory accesses for the output activations by caching partial sums on-chip. Nonetheless, \tool is adaptable to other dataflows with minimal modifications.

\textbf{The Data Movement Bottleneck:}
From a data movement perspective, this approach requires the transfer of {\small $\left\lceil \frac{P}{T_P}\right\rceil$} tiles of size {\small $T_R$$\times$$T_P$} for the inputs, {\small $\left\lceil \frac{P}{T_P}\right\rceil$} tiles of size {\small $T_P$$\times$$T_C$} for the weights, and one tile of size {\small $T_R$$\times$$T_C$} for the outputs. To produce all the output tiles, all the data movements are performed {\small $\left\lceil \frac{R}{T_R}\right\rceil \left\lceil \frac{C}{T_C}\right\rceil$} times. 
In spite of the compute-bound CONV layers, the external memory bandwidth often becomes the bottleneck in CNN inference. This is primarily manifested in cases where: \textit{i)}~a resource-rich FPGA device is targeted. In this case, a large and powerful processing engine is instantiated and the speed of feeding it with new data constrains the performance; \textit{ii)}~the CNN layers have a large amount of weights, either due to large kernel sizes or number of filters. This case often occurs in deeper CNN layers, which are typically of significant width. As such, the weights cannot be stored on-chip and multiple memory transactions have to be issued, putting pressure on the available bandwidth; \textit{iii)}~high-dimensional input and output activations have to be transferred, increasing excessively the bandwidth requirements. This typically occurs in earlier layers of a CNN, where the feature maps dimensions are still large. 

As such, there is an emerging need for relieving the data movement burden and address the impact of hitting the memory wall. In this context, \textit{on-the-fly weights generation} can be an enabling factor in extracting higher performance and making more cost-effective use of hardware CNN engines.



\subsection{Devising a Hardware CNN Weights Generator}

\begin{figure*}[t]{
    \includegraphics[width=0.8\textwidth]{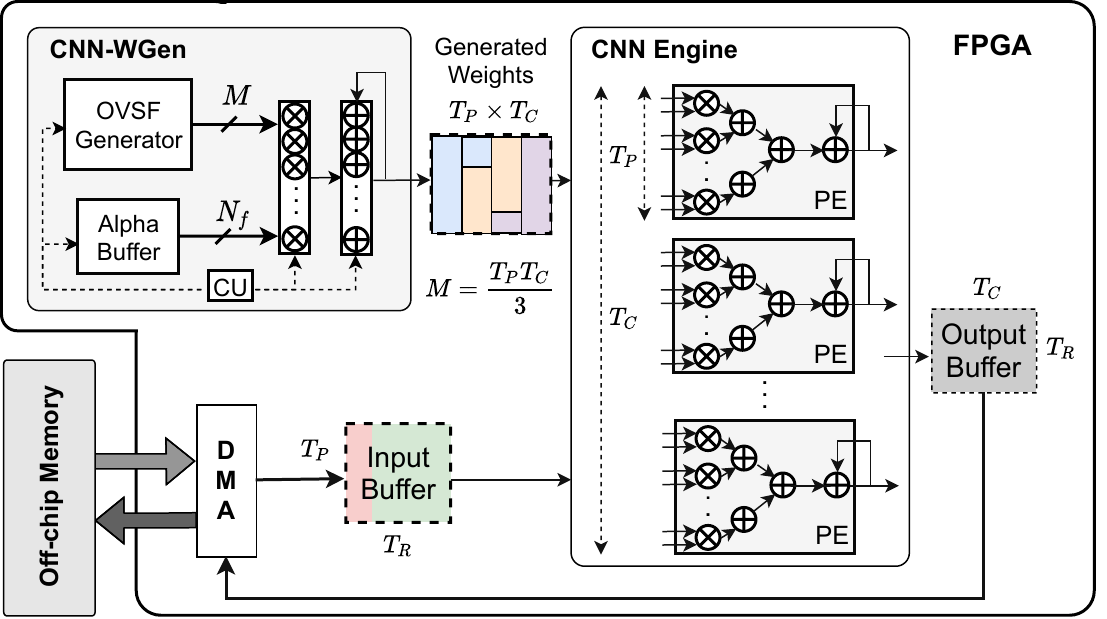}
    \vspace{-0.2cm}
}
\captionsetup{font=small,labelfont=bf}
\caption{CNN engine with on-the-fly weights generation. The weights are dynamically generated and the full bandwidth is available for inputs and outputs.}
\label{fig:unzipfpga_cnn_engine}

\end{figure*}

The objective is to minimise the data movement of CNNs by dynamically constructing the model weights using only on-chip resources. 
Importantly, on-the-fly weights generation needs to take place at run time, in a timely and per-layer fashion, since each CNN layer is a standalone independent schedulable unit. Moreover, the amount of computational and memory resources assigned to the weights generator ought to be balanced with the rest of the CNN engine to maximise the throughput of the accelerator while sustaining high resource utilisation. To that end, bringing forth on-the-fly weights generation requires devising two major components:


\subsubsection{Tiled Weights Generation}
Our novel insight is that, to be able to generate weights for layers of various dimensions, there is a need for a tiling method on top of the weights generation process. We denote our proposed tiling method by TiWGen. As shown in Fig.~\ref{fig:unzipfpga_cnn_engine}, TiWGen divides each $T_P$$\times$$T_C$ weights tile into subtiles of size $M$, with $M$ being uniform across the CNN's layers. Tiling on top of the weights generation method makes the dataflow of diverse layers identical to each other. With this approach, the value of $M$ becomes independent of the CNN architecture and is solely bound by the resources allocated to the weights generator. As such, $M$ exposes a tunable trade-off between weights generation speed and resource consumption.

\SetArgSty{textnormal} 
\setlength{\textfloatsep}{0pt}
\begin{algorithm}[!t]	
	\footnotesize
	\SetAlgoLined
	\LinesNumbered
	\DontPrintSemicolon
	
	\KwIn{Layer's weights matrix shape:  $P \times C = N_{\text{in}}K^2 \times N_{\text{out}}$}
	\nonl
	\myinput{Row and column tile sizes $T_P$ and $T_C$}
	\nonl
	\myinput{$\alpha$ values with $\alpha \in \mathbb{R}^{N_\text{in}N_\text{out}\left\lceil \rho K^2 \right\rceil}$}
	
	\KwOut{Weights matrix $\boldsymbol{W}$}
	
    \For(\Comment{\textit{tiles loop} - \textbf{\#} \textbf{\texttt{PIPELINE}}}){$t\gets1$ \KwTo $\left\lceil \frac{P}{T_P} \right\rceil$$\cdot$$\left\lceil \frac{C}{T_C} \right\rceil$}{
	   \For(\Comment{\textit{subtiles loop} - \textbf{\#} \textbf{\texttt{PIPELINE}}}){$i\gets1$ \KwTo $\left\lceil \frac{T_P T_C}{M} \right\rceil$}{
    	    $\text{subtile}^t_i \leftarrow \mathbf{0}$ \;
    	    \For(\Comment{\textit{basis vectors loop} - \textbf{\#} \textbf{\texttt{PIPELINE}}}){$j\gets1$ \KwTo $\rho K^2$}{
    	        \For(\Comment{\textbf{\#} \textbf{\texttt{UNROLL}}}){$k\gets1$ \KwTo $M$}{
    	            $incr_k \leftarrow \text{vec}_j(k) \cdot \alpha_k$ \hspace{0.37in}\Comment{\textit{Multiplier array}} \;
    	            $\text{subtile}^t_i(k) \leftarrow \text{subtile}^t_i(k) + incr_k$ \hfill\Comment{\textit{Adder array}}\;
    	        }
    	    }
    	    $\text{tile}^t \leftarrow \text{UpdateTile}(\text{tile}^t,\text{subtile}^t_i)$\;
    	}
    	$\boldsymbol{W} \leftarrow \text{UpdateMatrix}(\boldsymbol{W}, \text{tile}^t)$\;
	}

	\caption{\footnotesize 
	Generation of a layer's weights using TiWGen
	}
	\label{alg:tiled_weights_gen}	
\end{algorithm}

Alg.~\ref{alg:tiled_weights_gen} describes the internal workings of TiWGen. Initially, the $P$$\times$$C$ weights matrix is partitioned into {\small $\left\lceil \frac{P}{T_P}\right\rceil$} tiles of size $T_P$$\times$$T_C$, with each tile processed sequentially (line~1). Next, each tile is divided into $\left\lceil \frac{T_P T_C}{M} \right\rceil$ subtiles~(line 2). After all basis vectors of the current subtile have been processed (lines~4-9), the associated part of the output tile is updated (line~10) and the algorithm proceeds to the next subtile. When all subtiles of a tile have been generated, the weights matrix is updated (line~12) and the algorithm continues to the next iteration until all weights tiles have been constructed.  

\rev{\textbf{Applicability to Other Dataflows.}~Although the presented instance of TiWGen focuses on output stationary dataflows, our method can be applied to hardware designs that employ other dataflows. The main modifications comprise \textit{i)}~the order the generated weights and \textit{ii)}~the required generation rate. For instance, considering Google’s TPU~\cite{tpu2017isca} which is the most widely used systolic array for CNN inference, the accelerator adopts a weight-stationary dataflow. In this case, as the tile of the weight matrix is reused for several cycles, the OVSF generator would have to generate weights in longer periods compared to output stationary dataflow and the resource allocation would be automatically adjusted at the DSE stage accordingly.}

\subsubsection{Weights Generator Microarchitecture}
\begin{figure}[t]
    \centering
    \vspace{-0.1cm}
    {
    \includegraphics[width=0.7\columnwidth,trim={0.45cm 0cm 2.8cm 0cm},clip]{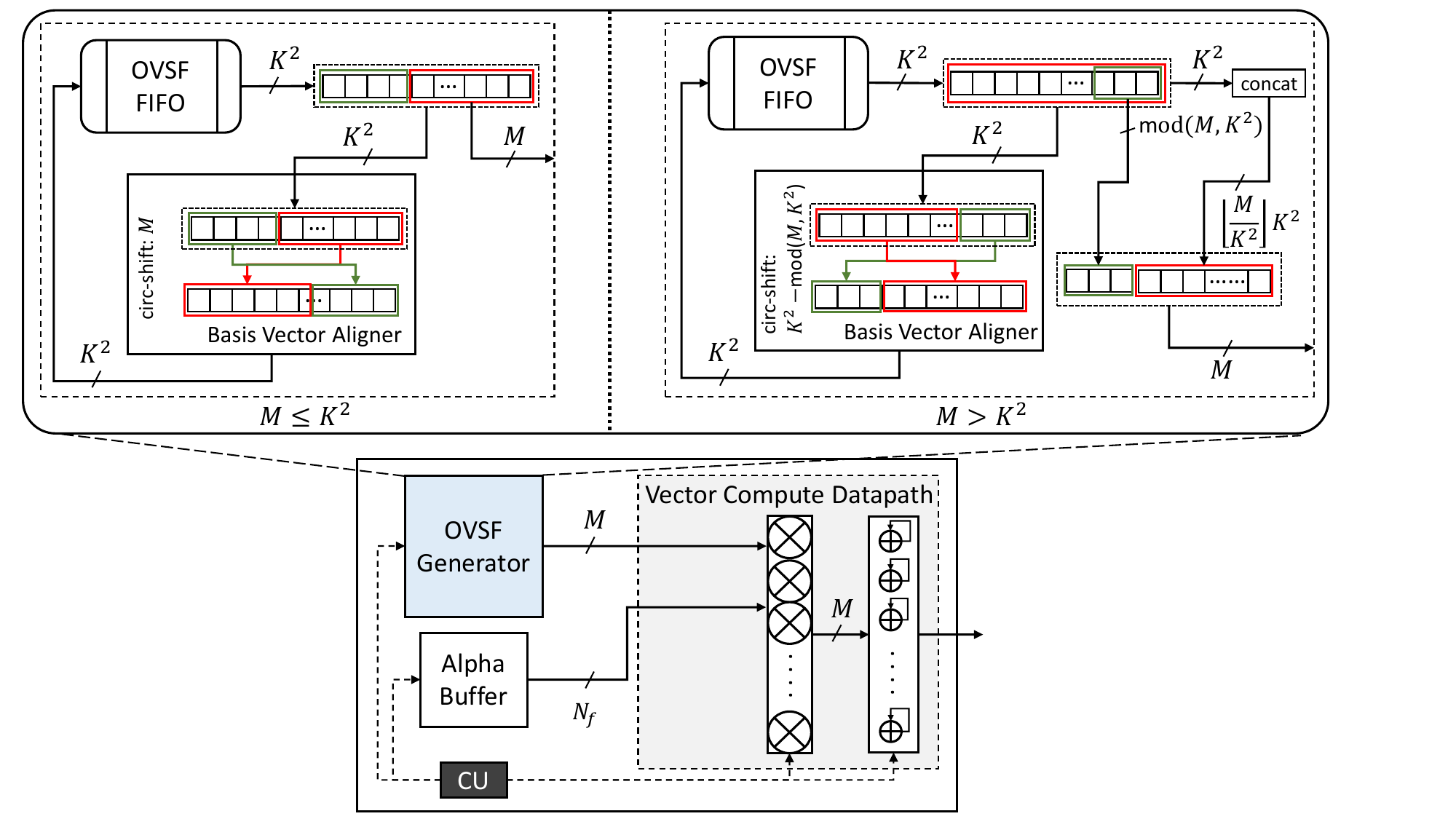}
    }
    \vspace{-0.4cm}
    \captionsetup{font=small,labelfont=bf}
    \caption{\footnotesize Microarchitecture of the \texttt{CNN-WGen} module.}
    \label{fig:hw_weights_gen}
\end{figure}

With the design objectives and constraints of Section~\ref{sec:design_flow} in mind, we propose a microarchitectural unit, called \texttt{CNN-WGen}, which is placed within the CNN engine (Fig.~\ref{fig:unzipfpga_cnn_engine}) and is responsible for generating the weights in an orderly manner and feeding them to the processing engine. Fig.~\ref{fig:hw_weights_gen} illustrates the design. As shown, the unit consists of: \textit{i)}~a \textit{vector compute datapath} comprising two vector units (multiplier and adder arrays), \textit{ii)}~the \textit{Alpha buffer} storing the $\alpha$ values, and \textit{iii)}~the \textit{OVSF generator} that is responsible for outputting the $M$-sized basis vector subtiles as dictated by the TiWGen scheme.

\textbf{Mapping Strategy.}
To efficiently map and perform the TiWGen loops, \texttt{CNN-WGen} employs loop optimisation techniques, annotated in Alg.~\ref{alg:tiled_weights_gen}. Namely, loop pipelining and unrolling are employed to customise the computation patterns and on-chip memory reuse of the weights generator. Pipelining is applied on the three outer loops over tiles (line~1), subtiles (line~2) and basis vectors (line~4), and unrolling on the inner loop that processes the $M$-sized subtile (line~5). To unroll the innermost loop, \texttt{CNN-WGen} employs two $M$-wide vector units that perform $M$-parallel multiplications and additions, respectively. In this manner, tuning $M$ can balance the parallelism-resource usage trade-off of the module.

\textbf{Parametrised Vector Compute Datapath.}
As shown in Fig.~\ref{fig:hw_weights_gen}, the vector arithmetic units must have a fixed size that complies with the resource constraints of the target FPGA and namely the available DSP blocks. The multiplier array is connected to the \textit{OVSF generator} and the \textit{Alpha buffer}. For the i-th subtile (line~3 in Alg.~\ref{alg:tiled_weights_gen}), the former produces $\rho K^2$ basis vectors of size $M$, while the latter outputs the associated $\alpha$ coefficients, both of which are forwarded to the multiplier array in a pipelined manner.
All $M$ elements are processed in parallel by the $M$-wide vector units, leading to the vectorised unrolling of the inner loop on line~5 of Alg.~\ref{alg:tiled_weights_gen}. The adder array processes the output of the multiplier array by accumulating the $\rho K^2$ partial results. Finally, when TiWGen proceeds to the next subtile (\textit{i.e.}~next iteration of the loop on line~2), the control unit (CU in Fig.~\ref{fig:hw_weights_gen}) resets the accumulators' state.
Overall, the vector compute datapath is design-time configurable with respect to parameter $M$ which controls the sizing of the vectors units and balances in this way the performance-resource usage trade-off of \texttt{CNN-WGen}. The design space exploration of $M$ is discussed in Section~\ref{sec:dse}.


\textbf{Memory Customisation in Alpha Buffer.}
TiWGen dictates that each subtile contains weights from $N_f$ distinct $K$$\times$$K$ filters. To sustain \texttt{CNN-WGen}'s throughput, an equal number of $\alpha$s have to be fetched in parallel from the \textit{Alpha buffer}. To accomplish this, we design the \textit{Alpha buffer} as a unified buffer with customised memory organisation and addressing.
Each layer contains $N_{\text{in}} N_{\text{out}} \left\lceil \rho_l K_l^2 \right\rceil$ distinct $\alpha$ values. 
As such, the \textit{Alpha buffer} is broken down to $N_P^{\text{Alpha}}$$=$$N_f$ independent multi-bank sub-buffers, with a depth of $D^{\text{Alpha}}$ (Eq.~(\ref{eq:alpha_buff_depth})) to accommodate $N_L$ layers.
%
\begin{equation}
    \footnotesize
    N_{f} = \left\lceil \frac{\min (T_P,M)}{K_\text{max}^2} \right\rceil \left\lfloor \frac{M}{T_P} \right\rfloor + \text{mod}(M,T_P) \left\lceil \frac{M}{K_\text{max}^2} \right\rceil
    \vspace{-1mm}
    \label{eq:filters_per_subtile}
\end{equation}
\begin{equation}
    \footnotesize
    D^{\text{Alpha}} = \overbrace{\sum_{l=1}^{N_L}}^{\text{for each layer}} \frac{\overbrace{N_{\text{in}}^l N_{\text{out}}^l \left\lceil \rho_l K_l^2 \right\rceil}^{\text{no. of $\alpha$ values}}}{N_P^{\text{Alpha}}} \quad \text{ ( \textit{Buffer depth} )}
    \label{eq:alpha_buff_depth}
    \vspace{-0.1cm}
\end{equation}
where $N_L$ is the number of layers, $N_{\{\text{in},\text{out}\}}^l$ the l-th layer's number of input/output channels and $\rho_l$ the compression ratio. Finally, the outputs of the sub-buffers are concatenated and connected to the multiplier array to provide concurrent access to $N_f$ coefficients. Finally, if the number of $\alpha$ coefficients exceeds the available on-chip memory, the remaining coefficients are transferred from the off-chip memory.

\textbf{Rate Matching in OVSF Generator.}
Following TiWGen, the basis vectors are processed in a blocked manner with a tile size of $M$. This approach leads to two pipelined loops over the $\left\lceil \frac{T_P T_C}{M} \right\rceil$ subtiles (line~2) and the $\rho K^2$ basis vectors (line~4) and the unrolled loop of processing the $M$-element subtile with the vector units (line~5). To produce the i-th subtile, the \textit{OVSF generator} feeds the compute datapath with $\rho K^2$ basis vectors that are tiled as dictated by TiWGen's parameter $M$.

In order not to straggle the operation of \texttt{CNN-WGen}, the \textit{OVSF generator} has to match the rate of the vector units by feeding them with $M$ bits/cycle. A conventional design involves statically laying out the tiled vectors into a single buffer, with $M$ ports and a depth equal to the number of reads per tile (\textit{i.e.}~\#basis vectors$\times$\#subtiles).
However, such a monolithic design would impose significant overheads as the basis vectors would have to be replicated either in the same address (\textit{e.g.}~when $M$$>$$K^2$) or in multiple addresses (\textit{e.g.}~storing rotated versions as required by different subtiles). This leads to inefficient utilisaiton of the on-chip memory due to excessive replication.

An alternative approach that would avoid the basis vector replication involves the instantiation of a $K^2$-deep OVSF memory with each location storing one $K^2$-bit vector. Such a design requires significantly lower amount of storage and provides an access rate of 1 vector/cycle by reading the appropriate address. Nonetheless, to obtain the $M$-bit subtile from the $K^2$-bit vectoc, complex multiplexer selection circuitry has to be instantiated. This approach can affect the maximum clock rate or add latency cycles to such an extent that any throughput gains would be outweighed.

To alleviate these limitations when mapping TiWGen's tiling scheme, a custom \textit{OVSF generator} was developed. The top-level diagram of the \textit{OVSF generator} is shown in Fig.~\ref{fig:hw_weights_gen}. It is composed of three main components: the \textit{OVSF FIFO}, a \textit{basis vector aligner} and the \textit{output register}. By introducing a FIFO for the OVSF vectors in combination with a \textit{basis vector aligner}, the \textit{OVSF generator} introduces a rate-matching mechanism that sustains the processing rate of the \textit{vector compute datapath} while efficiently utilising the on-chip memory. The generator performs a different operation depending on the values of $M$ and $K^2$ of layer $i$ (Fig.~\ref{fig:hw_weights_gen}).

Initially, the \textit{OVSF FIFO} stores the \mbox{{\small ($K_i^2K_i^2$)}-bit} basis vectors. 
The current vector is read from the FIFO into the top register. 
If {\small $M$$\le$$K_i^2$}, 
the $M$ least significant bits (LSBs) are outputted to the \textit{vector compute datapath}. At the same time, the basis vector is processed by the \textit{basis vector aligner}, which performs an $M$-bit left circular shift and writes the rotated vector to the \textit{OVSF FIFO}.
If {\small $M$$>$$K_i^2$}, 
the basis vector is self-concatenated {\small $\left\lfloor \frac{M}{K_i^2} \right\rfloor$} times and written to the output's LS part. Simultaneously, the {\small $\text{mod}(M,K_i^2)$} LSBs of the basis vector are written to the output's MSBs and the constructed vector is passed to the compute datapath. Here, the aligner performs a left circ-shift of \mbox{{\small $K_i^2$-$\text{mod}(M,K_i^2)$}} bits and writes the result to the FIFO.

With this approach, when the basis vectors are read again out of the OVSF FIFO after $\rho K_i^2$ cycles (\textit{i.e.}~in the next iteration of the loop on line~2), they are correctly aligned to directly match TiWGen's tiling pattern. For instance, after the generation of the red-striped subtile in Fig.~\ref{fig:unzipfpga_cnn_engine}, the FIFO-read basis vectors will be correctly aligned in order to generate the blue-striped subtile without the need for costly selection logic or redundant storage. 
For CNNs with multiple filter sizes, the \textit{basis vector aligner} is instantiated with as many circ-shift options. As the distinct filter sizes are known \textit{a priori}, only the required shifting logic is inserted, avoiding expensive generic multiplexers, and the appropriate per-layer bit-shift is selected at run time.

Overall, the proposed design offers two main benefits. First, it alleviates the redundant replicated storage of basis vectors and avoids the hardware cost of partitioning multiplexers that would require excessive LUTs usage. Second, it provides the necessary bandwidth to the \textit{vector compute datapath} while efficiently utilising the on-chip memory through the \textit{OVSF FIFO} and the resource-efficient aligner design.
As the values of $K_i^2$ for each layer and $M$ are known at design time and after the DSE phase respectively, the \textit{OVSF generator} can be statically instantiated at compile time. 

\begin{wrapfigure}{R}{0.45\textwidth}
    \centering
    \vspace{-0.6cm}
    {
    \includegraphics[width=0.45\columnwidth,trim={6.5cm 5cm 10.25cm 1cm},clip]{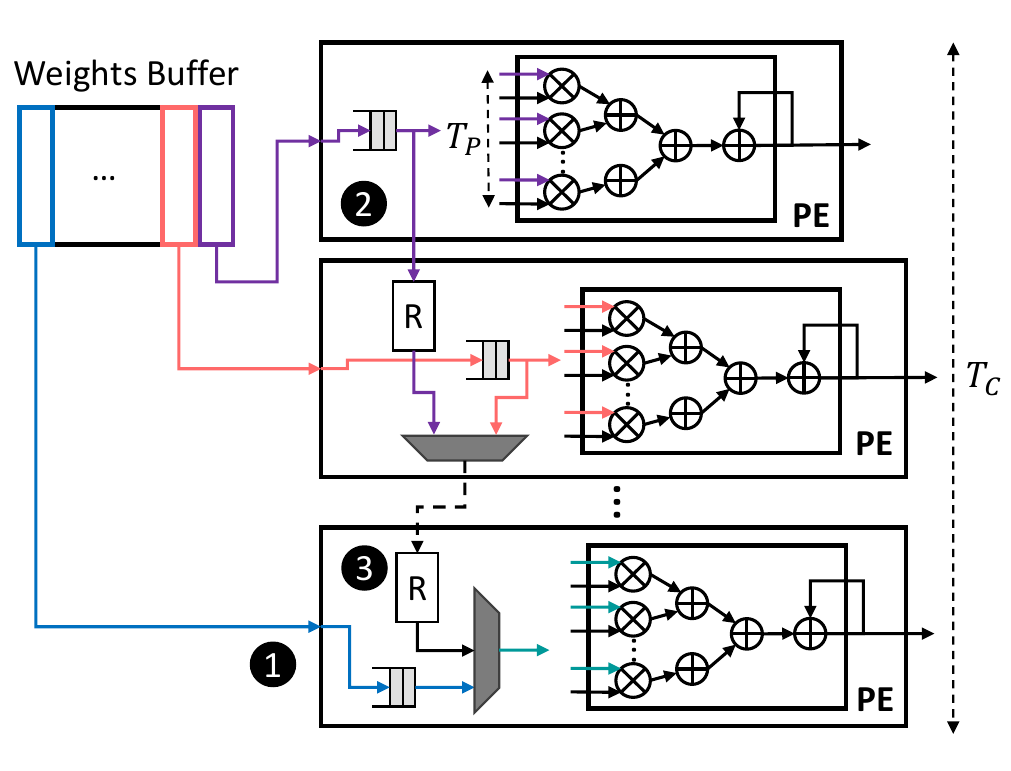}
    }
    \vspace{-0.6cm}
    \captionsetup{font=small,labelfont=bf}
    \caption{\footnotesize \tool's input selective PE array for CNNs.}
    \vspace{-0.2cm}
    \label{fig:new_pe_design}
\end{wrapfigure}

\vspace{0.4cm}
\subsection{Input Selective PEs for Counteracting Underutilisation}
\label{sec:pe_design}

One key limitation of existing CNN engines is that, when processing compute-bound layers, the layer dimensions often do not match the fixed processing engine configuration, leading to underutilisation of the computational resources and severe performance penalties~\cite{latency2017fpl,alamo2020tcad,caffeine2019tcad,maestro2020micro}.
Such a scenario can be observed when mapping a layer with $N_\text{out}$$=$$64$ channels (\textit{i.e.}~$C$$=$$64$) on an engine with 128 PEs (\textit{i.e.}~$T_C$$=$$128$). In this case, the PEs would remain idle 50\% of the time, halving the attainable performance.

To alleviate this, we propose \textit{input selective PEs}, a design that enables existing PEs to perform load-balancing through inter-PE work-stealing in a resource-efficient manner.
Fig.~\ref{fig:new_pe_design} shows \tool's input selective PEs. The initial PE is augmented with registers and switches. However, not all PEs have the same components; only the PEs that remain \textit{underutilised} even for a single layer are further equipped with a compact switch that selects the inputs to the dot-product circuit. 
In addition to the normal flow of data, these switches enable each PE to send its weight to its bottom neighbour. As highlighted in dark blue at the bottom PE of Fig.~\ref{fig:new_pe_design}, the switch on the left of the PE selects its input from \rev{two options: \textit{i)}~under normal operation, the PE is fed with the weight written by \texttt{CNN-WGen} in the weights buffer $\bigl(\protect\circled{1}\bigr)$; \textit{ii)}~in the absence of this weight (\textit{e.g.}~due to a mismatch between $C$ and $T_C$), the PE is fed with the weight passed by the adjacent PE $\bigl(\protect\circled{3}\bigr)$.}  
In the second case, the weights are propagated along the PE array \rev{$\bigl(\protect\circled{3}\bigr)$} so that a different weight is used by each augmented PE in each cycle. Moreover, the Input Buffer (Fig.~\ref{fig:unzipfpga_cnn_engine}) is reorganised accordingly to provide parallel access to multiple rows.

Effectively, this design works as a load-balancing mechanism that partially unrolls the $T_R$ dimension and thus distributes the work more evenly among the PEs. By restricting connectivity to adjacent units and enhancing only the underutilised PEs, the additional circuitry is low-overhead and delivers up to 20\% higher performance on compute-bound layers.

\section{Design Space Exploration}
\label{sec:dse}

Based on its parametrisation of the processing engine, buffer sizes and weights generator, \tool defines a particular architectural design space. To estimate the performance and resource usage of different configurations, an analytical modelling framework has been developed. 
At a high-level, the key decisions for yielding a high-performance configuration of the system are: the allocation of the on-chip resources between the CNN engine and the weights generator and, the sizes of the activations buffers.
The design-time tunable parameters comprise \textit{1)}~$M$ that determines the TiWGen's tile size and the size of \texttt{CNN-WGen}'s vector units, \textit{2)}~tile sizes $T_C$ and $T_P$ that determine the number of PEs and MACs per PE, respectively, and $T_R$ affecting the size of the activations buffers.

\subsection{Performance Model}
\label{sec:perf_model}

The workload of a CNN with $N_L$ layers is represented as a sequence of {\small $W_i$$=$$\left<R_i,P_i,C_i\right>$} \textit{workload tuples} with $i~\in~\{1,...,N_L\}$.
Given a design point {\small $\sigma$$=$$\left<M,T_R,T_P,T_C\right>$}, the \texttt{CNN-WGen}'s runtime for generating the i-th layer's weights required 
to compute a {\small $(T_R\times T_C)$} 
output tile is given by
\vspace{-0.1cm}
\begin{equation}
    \footnotesize
    \vspace{-1mm}
    t_{\texttt{CNN-WGen}}^i(\sigma, W_i) = \left\lfloor \rho \cdot l \right\rfloor \cdot \left\lceil \frac{T_P \cdot T_C}{M} \right\rceil \cdot \left\lceil \frac{P_i}{T_P} \right\rceil
    \vspace{-0.5mm}
    \label{eq:wgen_exec_time}
\end{equation}
where $\rho$ and $l$ are the OVSF ratio and basis length, respectively, and with one factor for each of the pipelined loops in Algorithm~\ref{alg:tiled_weights_gen}.
With $\alpha$ values transferred upfront and the OVSF method generating all weights on-chip, the off-chip memory transfers involve only the input/output activations
\begin{equation}
    \footnotesize
    t_{\text{mem in}}^i(\sigma, W_i) = \frac{T_R \cdot P \cdot WL}{bw_\text{in}}, \quad t_{\text{mem out}}^i(\sigma, W_i) = \frac{T_R \cdot T_C \cdot WL}{bw_\text{out}}
    \label{eq:transfer_times}
\end{equation}
where $WL$ is the adopted wordlength, and $bw_{\{\text{in},\text{out}\}}$ are the memory bandwidths for transferring inputs/outputs.

With $T_C$ and $T_P$ dimensions unrolled, computing an output tile 
requires the pipelined processing of $\frac{P_i}{T_P}$ tiles for each of the $T_R$ rows. Hence, the processing engine's runtime for each output tile is estimated as {\small $t_{\text{eng}}^i(\sigma, W_i) = T_R \left\lceil \frac{P_i}{T_P} \right\rceil$}. With the input selective PEs, the runtime is refined as
\begin{equation}
    \footnotesize
    t_{\text{eng}^*}^i (\sigma, W_i) = \left( T_C-C_i + \left\lceil \frac{T_R \cdot C_i - (T_C-C_i) \cdot (C_i+1)}{T_C} \right\rceil \right) \cdot \left\lceil \frac{P_i}{T_P} \right\rceil
    \label{eq:cnn_engine_exec_time_enhanced}
\end{equation}
where dimension $T_R$ is partially unrolled by processing rows of $T_R$ through the underutilised PEs.

Overall, the accelerator forms a pipeline of three coarse stages:
the concurrent input transfer and weights generation, the CNN engine processing and the output transfer. 
In this context, the initiation interval of the architecture is given by the maximum initiation interval of the three-stage pipeline, calculated as
%
\begin{equation}
    \footnotesize
    \vspace{1mm}
    II^i(\sigma, W_i) = \max\left( \max\left(t_{\text{mem in}}^i, t_{\texttt{CNN-WGen}}^i \right), t_{\text{eng}^*}^i, t_{\text{mem out}}^i \right)
    \label{eq:initiation_interval}
\end{equation}
As such, the total runtime for layer $i$ is given by 
\mbox{{\footnotesize $t_\text{total}^i(\sigma, W_i)=II^i(\sigma, W_i) \left\lceil \frac{R_i}{T_R} \right\rceil \left\lceil \frac{C_i}{T_C} \right\rceil$}}.
Thus, for a CNN with $N_L$ layers, the workload tuple is {\footnotesize $W$$=$$\left<W_i ~|~ \forall i \in \{1, ..., N_L \} \right>$} and the throughput in inferences per sec (inf/s) is estimated as \mbox{{\footnotesize $T(\sigma, W) = 1/\sum\limits_{i=1}^{N_L} t_\text{total}^i (\sigma, W_i)$}}.

\subsection{Resource Consumption Model}
The primary factor that constrains the mapping of a CNN engine on a given platform is resource availability. Each candidate configuration has a corresponding resource consumption. We define the \textit{feasible space} of our model as the set of configurations that satisfy all the platform-specific resource constraints. 
In our context, the main design constraints are the DSPs and on-chip RAM blocks of the target FPGA. Assuming that all MAC operators are mapped to DSPs, the values of {\small $
\left<M,T_P,T_C\right>$} are constrained as \mbox{{\footnotesize $D_\text{MAC} \times (M + T_P T_C) \leq D_\text{fpga}$}},
with {\footnotesize $D_{\text{fpga}}$} the available DSPs and {\footnotesize $D_\text{MAC}$} the DSPs/MAC. We consider 16-bit fixed-point precision, where {\footnotesize $D_\text{MAC}$$=$$1$} on the evaluated Xilinx FPGAs.


In terms of on-chip RAM, the accelerator has the I/O and Alpha buffers with wordlength $WL$ and the binary OVSF FIFO, with a total capacity requirement as given by Eq.~(\ref{eq:onchip_ram}).
%
\begin{equation}
    \vspace{1mm}
    \left(2(T_R T_P + T_R T_C) + D^{\text{Alpha}}N_P^{\text{Alpha}}\right) WL + K_\text{max}^2K_\text{max}^2 \leq C_\text{fpga}
    \label{eq:onchip_ram}
\end{equation}
where the factor of 2 accounts for double-buffering and $C_\text{fpga}$ is the on-chip RAM capacity of the target device.

To further estimate the consumption of look-up tables (LUTs), we used a set of place-and-route measurements and fitted linear regression models as a function of \tool's tunable parameters.
Overall, we formally capture the 
resource consumption of a design point $\sigma$ by means of vector $\textbf{\textit{rsc}}(\sigma)$ that holds the utilised amount of DSPs, BRAMs and LUTs. Similarly, we denote the FPGA resource vector by $\textbf{\textit{rsc}}_\text{Avail.}$. 

\subsection{Configuration Optimisation Framework}
To yield the highest performing design for the given CNN-FPGA pair, we cast the DSE task as a constrained optimisation problem that aims to determine the values of the configurable parameters $\left< M, T_R, T_P, T_C \right>$ that achieve the highest performance for the target CNN and available hardware resources. Formally, we express this setup as
\begin{equation}
    \footnotesize
    \min\limits_{\sigma=\left<M,T_R,T_P,T_C\right>} T\left(\sigma, W \right) \quad \text{s.t.} \quad \textbf{\textit{rsc}}(\sigma) \leq \textbf{\textit{rsc}}_\text{Avail.}
    \label{eq:opt}
\end{equation}
where $T$, $\textbf{\textit{rsc}}$ and $\textbf{\textit{rsc}}_{\text{Avail.}}$ are the throughput in inferences per second (inf/s), the resource consumption of the current design point $\sigma$ and the resource vector of the target platform, respectively.
Given a CNN-FPGA pair, we perform exhaustive search for different resource allocations between \texttt{CNN-WGen} and the processing engine. 
Designs that violate the resource constraints are pruned as infeasible to accelerate the exploration.

\section{Deriving Lightweight OVSF Models}
\label{sec:light_ovsf_models}

Having presented \tool's hardware architecture, its strategy for mapping OVSF models on the accelerator and its design space exploration process, we now describe important challenges for constructing efficient OVSF models. The main challenges comprise: \textit{i)}~extracting correctly-sized filters from OVSF codes, \textit{ii)}~selecting a subset of OVSF vectors to meet a given OVSF ratio for each layer, and \textit{iii)}~setting the per-layer OVSF ratios themselves. Section~\ref{sec:OVSF_issues} discusses our approach to \textit{i)} and \textit{ii)}, while Section~\ref{sec:hw_aware_ratios} introduces our novel hardware-aware scheme for tuning OVSF ratios.

\subsection{Practical Considerations to Train OVSF Models}
\label{sec:OVSF_issues}
Unlike standard CNNs, architectures using OVSF codes do not learn convolutional ﬁlters directly. Instead, they learn weighting coeﬃcients for each OVSF code representing a filter. However, despite their simplicity as a straight drop and replacement option for standard convolutional layers, the nature of OVSF codes and the filter generation process, present two fundamental challenges: \textit{1)}~OVSF codes are of power-of-two length: This constrains the generation of filters with all $N_{\text{out}}$, $N_{\text{in}}$, and $K$ being power-of-two integers. While this might be reasonable for the input and output channel dimensions, it prevents the construction of $3$$\times$$3$ filters, which are ubiquitous in modern CNN architectures; and \textit{2)}~choosing a subset of basis: Model compression is only achieved when OVSF ratio $\rho<$ $1$, which raises the question of \textit{which bases to choose} from the total $L$ available for OVSF codes of length $L$. Intuitively, their should be an optimal subset of basis for a given $\rho<$ $1$ that allows the learning of more expressive filters.

For \textit{1)}, we consider between \textit{i)}~utilising the first $\lfloor \rho \cdot K^2\rceil$ codes and \textit{ii)}~iteratively discarding OVSF codes based on their associated scalar $\alpha$ until the target compression ratio $\rho$ is reached. Compared to \textit{ii)}, with \textit{i)} we have a simpler optimization objective at the expense of potentially limiting the expressivity of OVSF filters. For \textit{2)}, we consider \textit{i)}~extracting a $3$$\times$$3$ crop from a $4$$\times$$4$ filter and \textit{ii)}~learning a mapping to a $3$$\times$$3$ filter by means of an average pooling layer. Similarly to the first pair of solutions, \textit{i)} represents a simpler training stage at the expense of a reduced effective field over the OVSF basis when constructing $3$$\times$$3$ filter. In Sec.~\ref{sec:basis_selection_and_3x3_extraction}, we compare both pairs of approaches for the above challenges.

In certain scenarios, a pre-trained model with standard convolutions might be available or can be trained very cheaply. In such cases, the formulation in Eq.~(\ref{eq:vector_eq}) could be reinterpreted as a minimisation problem and regress the set of $\pmb{\alpha}^*_i$ that minimise the difference w.r.t the standard filter $\hat{f}_{i}$ as \mbox{{\footnotesize$\pmb{\alpha}^{*}$$=$$\argmin_{\alpha}\left\|f - \hat{f} \right \|_2^2$}}, which can be implemented as a 2-layer MLP regression stage. We leverage this strategy when training OVSF models on ImageNet. More details are provided in Section \ref{sec:training_scheme}. 

\begin{figure*}[t]{
    \includegraphics[width=0.95\textwidth]{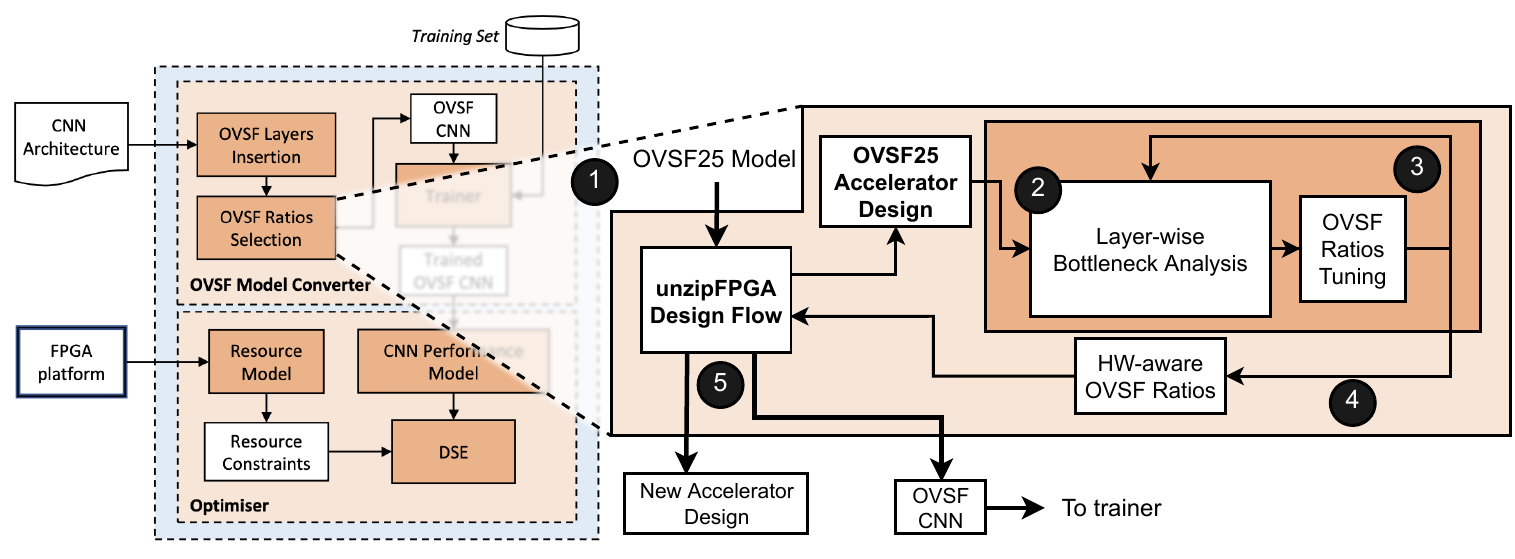}
    \vspace{-0.2cm}
    }
    \captionsetup{font=small,labelfont=bf}
    \caption{Overview of the proposed hardware-aware tuning of the OVSF ratios.}
    \label{fig:hw_aware_tune_flow}
\end{figure*}

\subsection{Hardware-Aware Tuning of OVSF Ratios}
\label{sec:hw_aware_ratios}

A critical component of \tool is the \textit{OVSF Ratios Selection} module of the \textit{OVSF Model Converter} (Fig.~\ref{fig:design_flow}).
In the original work~\cite{unzipfpga2021fccm}, the OVSF ratios (\textit{i.e.}~$\rho$ for each layer) were manually selected in a coarse, per-block manner, with the objective to reach a given compression ratio while minimising accuracy degradation. For example, as detailed in Section~\ref{sec:training_scheme}, to achieve a compression of 50\% in model size, denoted by OVSF50, the hand-tuned ratios were set by the tuple $[1.0, 0.5, 0.5, 0.5]$, indicating the OVSF ratio for each of the four blocks comprising a ResNet. Lower ratios were assigned to deeper layers as these contain a larger portion of the model parameters and are known to be more resilient to compression. The first CONV layer in the network remains untouched (\textit{i.e.}~not OVSF) as it has been shown to be less resilient to approximations including quantisation~\cite{alizadeh2018a}. To reach higher compression ratios with minimal accuracy drop, more involved tuning is required. This can be observed through the OVSF25 variant which achieves 75\% compression with diverse ratios of $[1.0, 0.4, 0.25, 0.125]$. Nonetheless, this process neither takes into account the impact of ratio values on hardware performance nor is automated, requiring elaborate tuning.

To alleviate this, we introduce a hardware-aware autotuning scheme for selecting the OVSF ratios of a given model on a target device. 
The key insight behind our method is that, for layers that are either compute- or memory-bound, we can allow the weights generation stage to consume more cycles by using more OVSF vectors (\textit{i.e.}~using a higher OVSF ratio) \textit{without} affecting the processing speed. As such, \texttt{CNN-WGen} will output a better approximation of the layer's weights, increasing the model's expressivity and potentially improving accuracy. With reference to our performance model (Section~\ref{sec:perf_model}), this case occurs for layer $i$ when either the processing engine's runtime ($t^i_{\text{eng}*}$) or the off-chip memory transfers ($t^i_{\text{mem in}}$ or $t^i_{\text{mem out}}$)  dominate the initiation interval in Eq.~(\ref{eq:initiation_interval}). This allows us to allocate more cycles for $t^i_{\texttt{CNN-WGen}}$ by using a higher OVSF ratio and obtaining a better approximation of the weights.

With this insight, Fig.~\ref{fig:hw_aware_tune_flow} presents our hardware-aware autotuning scheme is as follows. As a first step, we run \tool's design flow (Fig.~\ref{fig:design_flow}) using the OVSF25 ratios (\textit{e.g.}~$[1.0, 0.4, 0.25, 0.125]$ for ResNet) and derive the corresponding accelerator configuration $\bigl(\circled{1}\bigr)$. Next, we perform a bottleneck analysis of each layer's mapping on the accelerator that indicates which stage dominates the initiation interval $\bigl(\circled{2}\bigr)$, \textit{i.e.}~whether it is memory-bound (either input or output activation transfer), compute-bound or weights-generation-bound. For the layers where \texttt{CNN-WGen} is \textit{not} the bounding factor, we iteratively \stelios{Typically, we converged at 5 to 6 iterations in the examined models. We can report how much extra time that is, i.e. the tuning/search overhead - given the journal's theme, this might be expected. Else, we can leave it for the revised version after the reviews.} increase the OVSF ratios up to the point where the bottleneck does not shift to the weights generation stage $\bigl(\circled{3}\bigr)$. 
This leads to a more balanced pipelining of each layer, hence increasing accuracy by better utilising the instantiated accelerator. At the end of this process, the converged set of OVSF ratios are passed as the output of the \textit{OVSF Ratios Selection} module (Fig.~\ref{fig:design_flow}) and the rest of \tool's flow is run $\bigl(\circled{4}\bigr)$. As such, the model is retrained and the design space exploration is rerun with the new OVSF ratios $\bigl(\circled{5}\bigr)$, and the final model-accelerator pair are deployed on the target FPGA. Despite the additional retraining step, we note that the training protocol, \textit{i.e.}~all hyperparameters, remains the same throughout, without the need for further tuning.

\stelios{This paragraph is new.}
In step $\circled{3}$, the candidate values of ratio $\rho^l$ for the $l$-th layer lie in $\left\{ L^l/n ~|~ \forall \, n \in [1,L^l] \right\}$, where $L^l=N^l_{\text{in}}K^lK^l$ is the layer's code length (Sec.~\ref{sec:background_cnn_ovsf}). As such, there are $|L^l|$ candidate OVSF ratio values per OVSF layer. Overall, for a CNN with $N_L$ OVSF layers, there is a total of $\prod_{l=1}^{N_L} |L^l|$ possible OVSF ratio combinations. With an increase in either the model's depth or an OVSF layer's width, an enumerative exhaustive search quickly becomes computationally intractable. To alleviate this, we perform a parallel search for all OVSF layers, starting from the OVSF25 configuration.
At the first iteration, we first calculate the \textit{throughput ratio} between the weights-generation stage and the bottleneck stage of each layer and set the respective OVSF ratio to the closest feasible value. Then, we search the neighbouring candidate ratios until we find the maximum value that does not turn the weights-generation stage into the bottleneck. Throughout our experiments, this process lead to an average of 5 iterations to converge to the final hardware-aware set of ratios across the examined models.

Table~\ref{tab:hw_aware_tune_example} illustrates the impact of our scheme, denoted by \texttt{hw-aware-autotuning}, using ResNet18 on Z7045 for varying bandwidth availability. In the most bandwidth-constrained case (1.1~GB/s), OVSF25 is memory-bound, with all layers being limited by the transfer of the input feature maps. Our method exploits severe memory-boundedness and selectively increases the OVSF ratios, leading to an accuracy improvement of 1.2pp over OVSF25 with no sacrifice of the processing speed. For medium bandwidth levels (2.2~GB/s), a number of OVSF25 layers become compute-bound. If we naively set all OVSF ratios to 1.0 (shown as \texttt{uniform-1.0}), several layers become bound by the weights generation stage. Instead, with our bottleneck-guided method, the weights are more accurately generated while no change occurs to the boundedness of each layer. This results in achieving the same throughput as OVSF25, but with a 1.1pp increase in accuracy. Finally, in the high-bandwidth case (4.4~GB/s), our method introduces a 0.3pp accuracy gain, without affecting the hardware performance.
\stelios{Can comment more on the selection of ratios. Specifically, point out that our method picks the exact OVSF ratio values depending on the margin of each layer from changing bounding factor, e.g. there is room for higher ratios for some middle layers (0.333 in 4.4 GB/s) and less room in latter layers (0.25 ratios).}

Overall, our OVSF ratio selection method incorporates three features: \textit{1)}~it is fine-grained by allowing for different per-layer OVSF ratios within each block. As such, we obtain finer-grained control over the accuracy-compression trade-off; \textit{2)}~it bounds the accuracy drop from below by means of an informed initialisation of the ratio values. By starting from the OVSF25 ratios (\textit{i.e.} our most lightweight setting), we guarantee the accuracy's lowest bound and, by allowing only increases in OVSF ratios, we ensure that these would only potentially contribute accuracy gains; and \textit{3)}~it is hardware-aware as it is guided by the bottleneck analysis of each layer's processing.

\begin{table}[t]
    \normalsize
    \centering
    \captionsetup{font=small,labelfont=bf}
    \captionof{table}{\footnotesize Different OVSF ratio selection methods with respect to accuracy and bottleneck stage for ResNet18.}
    \vspace{-0.1cm}
    \resizebox{1.0\linewidth}{!}{
    \begin{tabular}{c|c|c|l|cccccccccccccccccccc}
   
        \toprule
         Memory & OVSF Ratio & Accuracy & & \multicolumn{20}{c}{Layer ID} \\
         Bandwidth & Selection Method & (\%) & & L0 & L1 & L2 & L3 & L4 & L5 & L6 & L7 & L8 & L9 & L10 & L11 & L12 & L13 & L14 & L15 & L16 & L17 & L18 & L19 \\ 
        \midrule

        \multirow{6}{*}{1.1 GB/s} & \multirow{2}{*}{OVSF25} & \multirow{2}{*}{67.3} & Bound & 
        IFM & IFM & IFM & IFM & IFM & IFM & IFM & IFM & IFM & IFM & IFM & IFM & IFM & IFM & IFM & IFM & IFM & IFM & IFM & IFM \\
        & & & OVSF Ratio & 1.0 & 1.0 & 1.0 & 1.0 & 1.0 & 0.4 & 0.4 & 1.0 & 0.4 & 0.4 & 0.25 & 0.25 & 1.0 & 0.25 & 0.25 & 0.125 & 0.125 & 1.0 & 0.125 & 0.125 \\
        
        \cline{3-24} & \multirow{2}{*}{\texttt{uniform-1.0}} & \multirow{2}{*}{N/A} & Bound & 
        IFM & IFM & IFM & IFM & IFM & IFM & IFM & IFM & IFM & IFM & IFM & IFM & IFM & IFM & IFM & IFM & IFM & IFM & IFM & IFM \\
        & & & OVSF Ratio & 
        1.0 & 1.0 & 1.0 & 1.0 & 1.0 & 1.0 & 1.0 & 1.0 & 1.0 & 1.0 & 1.0 & 1.0 & 1.0 & 1.0 & 1.0 & 1.0 & 1.0 & 1.0 & 1.0 & 1.0
        \\
        
        \cline{3-24} & \multirow{2}{*}{\texttt{hw-aware-autotuning}} & \multirow{2}{*}{68.5} & Bound & 
        IFM & IFM & IFM & IFM & IFM & IFM & IFM & IFM & IFM & IFM & IFM & IFM & IFM & IFM & IFM & IFM & IFM & IFM & IFM & IFM \\
        & & & OVSF Ratio & 
        1.0 & 1.0 & 1.0 & 1.0 & 1.0 & 1.0 & 1.0 & 1.0 & 1.0 & 1.0 & 1.0 & 0.5 & 0.5 & 0.5 & 0.5 & 0.5 & 0.5 & 0.5 & 0.5 & 0.25 \\
        \cline{2-24}

         \multirow{6}{*}{2.2 GB/s} & \multirow{2}{*}{OVSF25} & \multirow{2}{*}{67.3} & Bound & 
        IFM & IFM & IFM & IFM & IFM & C & C & IFM & C & C & C & C & IFM & C & C & C & C & IFM & C & C \\
        & & & OVSF Ratio & 1.0 & 1.0 & 1.0 & 1.0 & 1.0 & 0.4 & 0.4 & 1.0 & 0.4 & 0.4 & 0.25 & 0.25 & 1.0 & 0.25 & 0.25 & 0.125 & 0.125 & 1.0 & 0.125 & 0.125 \\
        
        \cline{3-24} & \multirow{2}{*}{\texttt{uniform-1.0}} & \multirow{2}{*}{N/A} & Bound & 
        IFM & IFM & IFM & IFM & IFM & C & C & IFM & C & C & C & C & IFM & C & C & C & W & IFM & W & W \\
        & & & OVSF Ratio & 
        1.0 & 1.0 & 1.0 & 1.0 & 1.0 & 1.0 & 1.0 & 1.0 & 1.0 & 1.0 & 1.0 & 1.0 & 1.0 & 1.0 & 1.0 & 1.0 & 1.0 & 1.0 & 1.0 & 1.0
        \\
        
        \cline{3-24} & \multirow{2}{*}{\texttt{hw-aware-autotuning}} & \multirow{2}{*}{68.4} & Bound & 
        IFM & IFM & IFM & IFM & IFM & C & C & IFM & C & C & C & C & IFM & C & C & C & C & IFM & C & C \\
        & & & OVSF Ratio & 
        1.0 & 1.0 & 1.0 & 1.0 & 1.0 & 1.0 & 1.0 & 1.0 & 1.0 & 1.0 & 0.5 & 0.5 & 0.5 & 0.5 & 0.5 & 0.5 & 0.5 & 0.5 & 0.5 & 0.25 \\
        \cline{2-24}

        \multirow{6}{*}{4.4 GB/s} & \multirow{2}{*}{OVSF25} & \multirow{2}{*}{67.3} & Bound & 
        IFM & C & C & C & C & C & C & OFM & C & C & C & C & IFM & C & C & C & C & IFM & C & C \\
         & & & OVSF Ratio & 1.0 & 1.0 & 1.0 & 1.0 & 1.0 & 0.4 & 0.4 & 1.0 & 0.4 & 0.4 & 0.25 & 0.25 & 1.0 & 0.25 & 0.25 & 0.125 & 0.125 & 1.0 & 0.125 & 0.125 \\
        
        \cline{3-24} & \multirow{2}{*}{\texttt{uniform-1.0}} & \multirow{2}{*}{N/A} & Bound & 
        IFM & C & C & C & C & C & C & OFM & C & C & W & W & IFM & W & W & W & W & IFM & W & W \\
        & & & OVSF Ratio & 
        1.0 & 1.0 & 1.0 & 1.0 & 1.0 & 1.0 & 1.0 & 1.0 & 1.0 & 1.0 & 1.0 & 1.0 & 1.0 & 1.0 & 1.0 & 1.0 & 1.0 & 1.0 & 1.0 & 1.0
        \\
        
        \cline{3-24} & \multirow{2}{*}{\texttt{hw-aware-autotuning}} & \multirow{2}{*}{67.6} & Bound & 
        IFM & C & C & C & C & C & C & OFM & C & C & C & C & IFM & C & C & C & C & IFM & C & C \\
        & & & OVSF Ratio & 
        1.0 & 1.0 & 1.0 & 1.0 & 1.0 & 1.0 & 1.0 & 1.0 & 1.0 & 0.333 & 0.333 & 0.5 & 0.333 & 0.333 & 0.333 & 0.25 & 0.25 & 0.25 & 0.25 & 0.25 \\

    \bottomrule
    \multicolumn{20}{l}{* IFM: Memory-bound w.r.t. input feature maps | OFM: Memory-bound w.r.t. output feature maps | C: Compute-bound | W: Weights Generation-bound.} \\
    \end{tabular}
    }
    
    \label{tab:hw_aware_tune_example}
\end{table}

\begin{table}[t]
	\centering
        \captionsetup{font=small,labelfont=bf}
	\caption{\footnotesize FPGA platforms used for evaluation.}
	\vspace{-0.2cm}
	\resizebox{0.65\linewidth}{!}{%
		\begin{tabular}{l l r l l r}
			\toprule
			\multicolumn{1}{l}{\multirow{1}{*}{Platform}} & Processor 
			& \multicolumn{1}{c}{\multirow{1}{*}{LUTs}} & Flip-Flops & DSPs & \multicolumn{1}{l}{\multirow{1}{*}{BRAM}} \\
			\midrule
			Zynq 7045 & Arm Cortex A9 
			& 218,600 & 437,200 & 900 & 2.40 MB \\
			UltraScale+ ZU7EV & Arm Cortex A53 
			& 230,000 & 461,000 & 1,728 & 4.75 MB \\
			\bottomrule
		\end{tabular}%
	}
	\label{tab:fpgas}
    \vspace{0.2cm}
\end{table}

\section{Evaluation}
\label{sec:eval}

\subsection{Experimental Setup}
\label{sec:exp_setup}

In our experiments, we target two widely used FPGA platforms with varied computational capabilities and memory resources (Table~\ref{tab:fpgas}): the Xilinx ZC706 board mounting the mid-tier Zynq Z7045 and the Xilinx ZCU104 board with the more resource-rich Zynq UltraScale+ ZU7EV. The two platforms are based on the Xilinx Zynq-7000 SoC and UltraScale+ MPSoC architectures, respectively, integrating a dual-core Arm Cortex A9 CPU and a quad-core Arm Cortex A53 CPU, respectively, alongside an FPGA fabric on the same chip. Our hardware designs were synthesised and placed-and-routed with Xilinx Vivado HLS and Vivado Design Suite (v2019.2) and run on both boards, with operating clock frequencies of 150~MHz for ZC706 and 200~MHz for ZCU104, respectively. The achieved clock frequency is currently constrained by the technology of the target device and the use of HLS, which relies on the vendor's toolchain and does not allow for low-level optimisations to shorten the critical path.

The corresponding Arm CPU was used to set up the transactions with the off-chip memory, launch the execution of inference and measure the end-to-end performance of each design. \tool provides support for both custom fixed-point and floating-point precisions. For the evaluation, 16-bit fixed-point precision was used, following the practice of the FPGA works we compare with. The available off-chip memory bandwidth was controlled by using a different number of memory ports and amount of word packing, spanning from 1.1 GB/s (1$\times$) to 13.4 GB/s (12$\times$).

\subsubsection{\textbf{Benchmarks}} 
\label{sec:benchmarks}
We evaluate on CNNs of varying depth, workload and memory footprint. Each CNN has been selected to impose a different design challenge. In particular, we target the widely used family of residual networks~\cite{DBLP:journals/corr/HeZRS15} and map variants of different depths to evaluate the scalability of our design. Concretely, we use ResNet18, ResNet34 and ResNet50 on the ImageNet dataset. In addition to image classification, ResNet models are also found as backbone of other tasks including object detection~\cite{detector2019iros}, super-resolution~\cite{mobisr2019mobicom} and semantic segmentation~\cite{deeplab2018tpami}. We also target SqueezeNet1.1~\cite{iandola2016squeezenet}, to assess \tool's efficacy on a highly optimised network for resource-constrained devices.

\subsubsection{\textbf{Basis Selection and \bm{$3$$\times$$3$} extraction}}
\label{sec:basis_selection_and_3x3_extraction}
The proposed on-the-fly formulation using OVSF codes allows for different strategies for: \textit{i)}~selecting which basis to use when $\rho$$<$$1$; and, \textit{ii)}~extracting $3$$\times$$3$ filters from \emph{true} OVSF filters that are restricted to be of shape $K$$\times$$K$ with $K$ being a power-of-two. In ~\ref{sec:OVSF_issues} we presented two solutions for each of aforementioned considerations. 
Table~\ref{tab:ovsf_cifar10} shows our analysis of the different approaches on CIFAR-10 with ResNet18/34. For the basis selection strategy \textit{i)}, iteratively dropping bases consistently yields higher-accuracy models. For \textit{ii)}, as the models become more compact (\textit{e.g.}~for OVSF50/25), cropping achieves higher accuracy compared to the average pooling approach.  
Thus, we leverage these findings to inform the parametrisation for ImageNet for the rest of the evaluation.

\subsubsection{\textbf{Training Scheme}}
\label{sec:training_scheme}
We have developed \tool's offline flow on top of \textit{PyTorch} (1.5). To derive the OVSF models, we modified the official \textit{PyTorch}-based ResNet by replacing all $3$$\times$$3$ convolutional layers within residual blocks with their OVSF counterparts. In all our experiments, we employed pre-trained ImageNet models from \textit{torchvision} (0.6.0). After a regression stage that transforms standard models into OVSF ones, the models were fine-tuned for 30 epochs using an Adam optimiser~\cite{kingma2014adam} and learning rate decay every 10 epochs. For each given model, we trained two OVSF variants following different distributions of ratios $\rho$ for layers in each of the four residual blocks. First, OVSF50 with ratios=$[1.0,0.5,0.5,0.5]$; and OVSF25 with ratios=$[1.0,0.4,0.25,0.125]$. 
We follow the same procedure and ratios for SqueezeNet's \textit{Fire} modules.

\subsubsection{\textbf{Baselines}}
\label{sec:baselines}

We introduce two highly optimised single computation engines executing: \textit{a)}~the vanilla CNN and \textit{b)}~pruned variants. For \textit{b)}, we use a state-of-the-art method~\cite{Molchanov_2019} which applies channel pruning based on the first-order Taylor approximation contribution of each filter to the model's loss. This process is carried out iteratively until a target compression ratio is reached. We refer to a pruned model that keeps 82\% of the filters as Tay82 and follow the same naming scheme for other ratios.
The baseline architecture comprises the conventional CNN engine design shown in Fig.~\ref{fig:conventional_cnn_engine}, with the weights transferred from the off-chip memory into the {\small $T_P$$\times$$T_C$} weights buffer, if they do not fit on-chip. Both \textit{a)} and \textit{b)} are parametrised with tile sizes {\small $\left<T_R,T_P,T_C \right>$} and roofline modelling~\cite{cnnroofline2015fpga} is used to obtain the highest throughput configuration for the target \mbox{CNN-FPGA pair}.

\subsection{Performance Comparison}
\label{sec:perf_comparison}
This section analyses the performance of the proposed framework with respect to both our optimised baselines and existing FPGA work.

\begin{table}[t]
    \normalsize
    \centering
    \captionsetup{font=small,labelfont=bf}
    \captionof{table}{\footnotesize Impact on accuracy for i)~each basis selection strategy and ii)~method to extract $3$$\times$$3$ filters from $4$$\times$$4$ OVSF filters. Models trained on CIFAR-10, with ResNet18/34 adapted for this dataset adn the much smaller variants ($\dagger$) proposed in~\cite{DBLP:journals/corr/HeZRS15}. Performing an iterative drop of bases, as opposed to taking the first $\lfloor \rho \cdot K^2\rceil$, consistently results in better models. As model size is reduced, taking a $3$$\times$$3$ crop from a $4$$\times$$4$ filter performed better than using an average pooling stage.}
    \vspace{-0.1cm}
    \resizebox{0.65\linewidth}{!}{
    \begin{tabular}{c|c|c|cc|cc|cccc}
   
        \toprule
        Model Arch. & Basis & Filters & \multicolumn{2}{c|}{OVSF100} & \multicolumn{2}{c|}{OVSF50} & \multicolumn{2}{c}{OVSF25}\\
        (baseline) & Strategy & to $3$$\times$$3$ & Param. & Acc. & Param. & Acc. & Param. & Acc.\\ \midrule
        
        \multirow{4}{*}{\parbox{1.9cm}{\centering ResNet18 93.2\% 11.2M}} & \multirow{2}{*}{Sequential} & Crop & \multirow{4}{*}{19.7} & 93.9 &  \multirow{4}{*}{9.1} & 93.7 & \multirow{4}{*}{3.6} & 92.9 \\
                                   & & Adaptive &  & 93.7 & & 93.8 & & 93.0 \\ \cline{2-3} \cline{5-5} \cline{7-7} \cline{9-9}
                                   & \multirow{2}{*}{Iterative} & Crop &  & 94.1 & & 93.6 & & \textbf{93.6} \\
                                   & & Adaptive &  & 94.0 & & 93.8 & & 92.3 \\ \midrule \midrule
                                   
        \multirow{4}{*}{\parbox{1.9cm}{\centering ResNet18$^\dagger$ 91.3\% 0.27M}} & \multirow{2}{*}{Sequential} & Crop & \multirow{4}{*}{0.48} & 90.8 & \multirow{4}{*}{0.25} & 90.8 & \multirow{4}{*}{0.15} & 88.3 \\
                                  & & Adaptive &  & 91.1 & & 91.2 & & 88.5\\ \cline{2-3} \cline{5-5} \cline{7-7} \cline{9-9}
                                  & \multirow{2}{*}{Iterative} & Crop &  & 91.1 & & 91.3 & & \textbf{91.4}\\
                                  & & Adaptive &  & 91.2 & & 91.4 & & 91.0\\ \midrule \midrule
                                   
        \multirow{4}{*}{\parbox{1.9cm}{\centering ResNet34 93.9\% 21.3M }} & \multirow{2}{*}{Sequential} & Crop & \multirow{4}{*}{37.7} & 94.1 &  \multirow{4}{*}{17.6} & 93.9 & \multirow{4}{*}{7.2} & 93.4  \\
                                  & & Adaptive &  & 94.3& & 94.0 & & 93.4 \\ \cline{2-3} \cline{5-5} \cline{7-7} \cline{9-9}
                                  & \multirow{2}{*}{Iterative} & Crop &  & 94.1& & 93.8 & & \textbf{94.3}\\
                                  & & Adaptive &  & 93.8 & & 93.7 & & 93.2\\ \midrule \midrule

        \multirow{4}{*}{\parbox{1.9cm}{\centering ResNet34$^\dagger$ 92.1\% 0.46M}} & \multirow{2}{*}{Sequential} & Crop & \multirow{4}{*}{0.82} & 92.3 & \multirow{4}{*}{0.43} & 91.4 & \multirow{4}{*}{0.26} & 89.3 \\
                                  & & Adaptive &  & 92.2 & & 91.5 & & 89.2\\ \cline{2-3} \cline{5-5} \cline{7-7} \cline{9-9}
                                  & \multirow{2}{*}{Iterative} & Crop &  & 92.3& & 91.8 & & \textbf{92.2}\\
                                  & & Adaptive &  & 92.4 & & 91.7 & & 91.7\\
                                   
    \bottomrule
    \end{tabular}
    }
    
    \label{tab:ovsf_cifar10}
\end{table}

\begin{table}[t]
    \tablefontsize
    \centering
    \captionsetup{font=small,labelfont=bf}
    \caption{\footnotesize Accuracy and number of parameters for ResNet34 models on ImageNet following different compression schemes. Performance measured on ZC706 at different memory bandwidths.}
    \vspace{-0.1cm}
    \resizebox{0.65\columnwidth}{!}{
    \begin{tabular}{l c c c c}
        \toprule
        Model & Compression & Params & Accuracy & Performance (inf/sec) \\
        Arch. & Method & (millions) & (\%) & ($1\times$, $2\times$, $4\times$) \\
        \midrule
        ResNet34 & - & 21.8 & 73.3 & (8.6, 16.8, 28.7) \\ 
        \midrule
        ResNet34 & Tay82 & 17.4 & $72.7$ & (10.7, 21.0, 35.6) \\ 
        ResNet34 & Tay72 & 15.1 & $71.9$ & (13.3, 25.8, 44.0) \\ 
        ResNet34 & Tay56 & 9.4 & $67.8$ & (18.3, 36.3, 63.8) \\ 
        ResNet34 & Tay45 & 6.3 & $63.1$ & (21.8, 43.4, 79.8) \\ 
        \midrule
        ResNet34 & OVSF50 & 17.2 & $72.8$ & (18.1, 21.8, 31.1) \\ 
        ResNet34 & OVSF25 & 7.2 & $71.5$ & (18.4, 27.3, 33.5) \\ 
        \midrule
        ResNet34 & Tay82+OVSF50 & 13.2 & $71.1$ & (18.6, 30.0, 37.3) \\ 
        ResNet34 & Tay82+OVSF25 & 6.7 & $70.6$ & (18.8, 31.0, 38.9 )\\ 
        ResNet34 & Tay72+OVSF50 & 11.9 & $70.3$ & (18.8, 32.0, 40.2) \\ 
        ResNet34 & Tay72+OVSF25 & 4.9 & $68.9$ & (18.9, 33.3, 42.0) \\ 
        \bottomrule
    \end{tabular}
    }
    \label{tab:accResultsResnet34}
\end{table}

\begin{table}[t]
    \tablefontsize
    \centering
    \captionsetup{font=small,labelfont=bf}
    \caption{\footnotesize Accuracy and number of parameters for ResNet18 models on ImageNet following different compression schemes. Performance measured on ZC706 at different memory bandwidths.}
    \vspace{-0.1cm}
    \resizebox{0.65\columnwidth}{!}{
    \begin{tabular}{l c c c c}
        \toprule
        Model & Compression & Params & Accuracy & Performance (inf/sec) \\
        Arch. & Method & (millions) & (\%) & ($1\times$, $2\times$, $4\times$) \\ 
        \midrule
        ResNet18 & - & 11.7 & 69.8  & (12.0, 23.5, 40.1)\\ 
        \midrule
        ResNet18 & Tay88 & 9.1 & $68.8$ & (14.3, 28.0, 46.4)\\ 
        ResNet18 & Tay82 & 7.9 & $67.3$ & (14.3, 27.8, 45.4)\\ 
        ResNet18 & Tay72 & 6.0 & $64.8$ & (18.2, 35.3, 57.6)\\ 
        ResNet18 & Tay56 & 3.7 & $58.3$ & (23.8, 47.3, 82.2)\\ 
        \midrule
        ResNet18 & OVSF50 & 9.1 & $69.2$ & (19.4, 33.8, 49.9)\\ 
        ResNet18 & OVSF25 & 4.1 & $67.3$ & (19.4, 34.8, 51.0)\\ 
        \midrule
        ResNet18 & Tay82+OVSF50 & 6.3 & $66.2$ & (24.5, 43.2, 57.9)\\ 
        ResNet18 & Tay82+OVSF25 & 2.8 & $64.4$ & (24.5, 43.6, 59.7)\\ 
        \bottomrule
    \end{tabular}
    }
    \label{tab:accResultsResnet18}
\end{table}

\subsubsection{\textbf{Comparison with Optimised Baselines}}
\label{sec:baseline_comparison}
Tables~\ref{tab:accResultsResnet34} and~\ref{tab:accResultsResnet18} show the achieved validation set accuracy and actual performance of each design as measured on ZC706 under varying bandwidth budget. Across bandwidths (1$\times$/2$\times$/4$\times$ where 4$\times$ is the 4.5 GB/s peak measured bandwidth on ZC706), \tool's OVSF50 and OVSF25 designs outperform the faithful baseline by {\small 2.1$\times$/1.3$\times$/1.1$\times$} and {\small 2.1$\times$/1.6$\times$/1.2$\times$} respectively for ResNet34, and by {\small 1.6$\times$/1.6$\times$/1.24$\times$} and {\small 1.4$\times$/1.5$\times$/1.3$\times$} respectively for ResNet18. As bandwidth availability increases, the baseline becomes less memory-bound and the performance gap closes. 
Table~\ref{tab:accResultsSqueezenet} shows the comparison of \tool with the faithful baseline for SqueezeNet on ZU7EV with peak measured bandwidth of 13.4~GB/s (12$\times$). Both OVSF50 and OVSF25 designs yield increasing throughput gains as the bandwidth becomes more restricted, with OVSF25 sustaining over 57\% speedup for up to 4$\times$ bandwidth.
Under 1$\times$ bandwidth, OVSF25 offers minimal additional gains. This is because, below a compression ratio, even though the memory needs are further reduced, activations begin to dominate I/O, and hence further weights reduction does not provide significant benefits. Activations compression techniques~\cite{eyeriss2017jssc,scnn2017isca} can be orthogonally combined to obtain further gains.

\rev{Based on our evaluation using SqueezeNet, we observe that computation can take place fast due to its lighter workload. As such, the attainable performance depends on how rapidly we can feed the CNN Engine with new inputs. Specifically, for the 4$\times$ bandwidth configuration, all layers of SqueezeNet are memory-bound. On the other hand, at 12$\times$ bandwidth, 88\% of the layers become compute-bound. As such, when there is restricted or medium availability of memory bandwidth, \tool significantly improves performance through our weights generation approach, with 78\%, 74\% and 55\% higher throughput for the 1$\times$, 2$\times$ and 4$\times$ bandwidth configurations, respectively (Table~\ref{tab:accResultsSqueezenet}). This improvement gradually decreases as the available bandwidth increases, with 15\% gain at 12$\times$ bandwidth.}

\begin{table}[t]
    \vspace{-0.2cm}
    \small
    \centering
    \captionsetup{font=small,labelfont=bf}
    \caption{\footnotesize Comparing \tool with faithful baseline on SqueezeNet on ImageNet. Performance measured on the UltraScale+ ZCU104 platform at different memory bandwidths.}
    \vspace{-0.1cm}
    \resizebox{0.65\linewidth}{!}{
    \begin{tabular}{l c c c c}
        \toprule
        Model & Compression & Params & Accuracy & Performance (inf/sec) \\
        Arch. & Method & (millions) & (\%) & ($1\times$, $2\times$, $4\times$, $12\times$)\\
        \midrule
        SqueezeNet & - & 1.24 & 58.2 & (72.9, 145.2, 290.4, 687.4) \\
        \midrule
        SqueezeNet & OVSF50 & 1.07 & 57.6 & (129.8, 252.9, 452.1, 792.1) \\
        SqueezeNet & OVSF25 & 0.86 & 57.1 & (129.8, 252.9, 456.8, 800.6) \\
        \bottomrule
    \end{tabular}
    }
    \vspace{0.2cm}
    \label{tab:accResultsSqueezenet}
\end{table}

\textbf{Comparison with Pruned Baselines.} Compared to the pruned baselines, \tool's OVSF models are more resilient at high compression ratios while resulting in similar accuracy at lower compression ratios. Informed by the analysis in Table~\ref{tab:ovsf_cifar10}, OVSF models are trainined to extract a $3$$\times$$3$ from a $4$$\times$$4$ and, to iteratively discard OVSF basis until the target compression ratio $\rho$ for each layer is reached, as first discussed in Sec.~\ref{sec:basis_selection_and_3x3_extraction}. In terms of throughput, \tool delivers faster processing at more constrained bandwidths. Concretely, ResNet34-OVSF50 is 80\% faster than Tay82 at 1$\times$ bandwidth, with less than 1 percentage point (pp) accuracy drop. Despite being almost identical in terms of model size and accuracy, Tay82's approach, which prioritises the pruning of layers with the least accuracy impact, leads to the pruning of mostly compute-bound layers when targeting ResNet34. On the other hand, ResNet34-OVSF50 compresses more effectively memory-bound layers, leading to significantly higher throughput at low bandwidths.
A similar pattern is observed for ResNet18. At higher compression ratios, ResNet34-OVSF25 yields 3.7 pp higher accuracy than Tay56, despite using 25\% fewer parameters.

To explore the benefits of combining \tool's OVSF execution scheme with channel pruning, we derive, train and map on \tool Tay-OVSF models. 
This results in competitive lightweight models that are not attainable through pruning alone. For instance, ResNet18 with Tay82+OVSF25 is 25\% smaller than ResNet18-Tay56 and achieves 6.1 pp higher accuracy, while achieving 34.6\% and 23.5\% higher throughput over ResNet18-Tay72 with less than 0.5 pp accuracy drop.

\begin{table*}[t]
	\centering
	\captionsetup{font=small,labelfont=bf}
    \caption{\footnotesize Comparison with prior FPGA work on ResNet18 (4.03 GOps), ResNet34 (7.40 GOps) \& SqueezeNet (0.78 GOps).}
	\vspace{-0.2cm}
	\resizebox{1.0\linewidth}{!}{
	\setlength\tabcolsep{10pt} 
		\begin{threeparttable}
			\small
			\begin{tabular}{@{}l l l| l l| l l l l@{}}
				\toprule
				Comparison with: 
				& \multicolumn{2}{l|}{\textbf{Compiler-based Design}}
				& \multicolumn{2}{l|}{\textbf{Compression-based Design}} 
				& \multicolumn{1}{l}{\textbf{Light-CNN-tailored Design}}
				& \multicolumn{2}{l}{\textbf{Multi-Accelerator Designs}}
				\\
				& ResNet18~\cite{snowflake2017compiling} 
				& \begin{tabular}[l]{@{}l@{}} \tool: \\ ResNet18* \end{tabular}
				& \begin{tabular}[l]{@{}l@{}} Sparse ResNet34~\cite{sparsecnnaccel2019fccm} \\ using Deep Compression  \end{tabular}
				& \begin{tabular}[l]{@{}l@{}} \tool: \\ ResNet34* \end{tabular} 
				& SqueezeNet~\cite{lightopu2020fpga} 
				& \multicolumn{2}{l}{SqueezeNet~\cite{maximising2017isca}} 
				& \begin{tabular}[l]{@{}l@{}} \tool:\\ SqueezeNet* \end{tabular}
				\\
				\cmidrule{7-8}
				\midrule
				FPGA  
				& Z7045 
				& Z7045 
				& Z7045 
				& Z7045
				& K325T
				& V485T 
				& V690T
				& ZU7EV 
				\\
				Clock (MHz) & 250 & 150 & 166	& 150 & 200 & 170 & 170 & 200 \\
				Precision & 16b fixed & 16b fixed & 16b fixed & 16b fixed & 8b fixed & 16b fixed & 16b fixed & 16b fixed \\
				DSPs$^\dagger$ & 900 & 900 & 900 & 900 & 840 & 2800 & 3600 & 1728 \\
				Logic Capacity & 218.6 kLUTs & 218.6 kLUTs & 218.6 kLUTs & 218.6 kLUTs & 203.8 kLUTs & 303.6 kLUTs & 433.2 kLUTs & 230.0 kLUTs \\
				Block RAM & 2.40 MB & 2.40 MB & 2.40 MB & 2.40 MB & 1.95 MB & 4.52 MB & 6.46 MB & 4.75 MB \\
				{\color{black}DSP Util.$^\dagger$} & 28.4\% & 100\% & 56.8\% & 100\% & 83.8\% & 80\% & 80\% & 100\% \\
				
				\begin{tabular}[t]{@{}l@{}} 
					inf/s
				\end{tabular} 
				& 21.38
				& 49.90
				& 27.84
				& 31.1
				& 420.90
				& 913.40
				& 1173.00
				& 792.20
				\\
				\begin{tabular}[t]{@{}l@{}} 
					inf/s/DSP$^\dagger$
				\end{tabular} 
				& 0.0237
				& 0.0576
				& 0.0309
				& 0.0369
				& 0.2505
				& 0.3260
				& 0.3258
				& 0.4584
				\\
				\begin{tabular}[t]{@{}l@{}} 
					inf/s/Logic
				\end{tabular} 
				& 0.0978
				& 0.2282 
				& 0.1273
				& 0.1422 
				& 2.0652
				& 3.0085
				& 2.7077
				& 3.444
				\\
			
				\bottomrule
				
			\end{tabular} 
			\begin{tablenotes}
				\small
				\item * using OVSF50, ** batch size = 1, $\dagger$ 18$\times$18, 19$\times$18 and 25$\times$18 DSP configurations, inf/s/DSP is adjusted based on precision for fair comparison (multiplied by 0.5 for 8b).
			\end{tablenotes}
			
		\end{threeparttable}
	}
	\vspace{-0.25cm}
	\label{tab:comparison_table}
\end{table*}

\begin{table*}[t]
	\centering
	\captionsetup{font=small,labelfont=bf}
    \caption{\footnotesize Comparison with prior FPGA work on ResNet50 (8.41 GOps).}
	\vspace{-0.2cm}
	\resizebox{1.0\linewidth}{!}{
	\setlength\tabcolsep{2pt} 
		\begin{threeparttable}
			\small
			\begin{tabular}{l l l | l l l l l l l l l l}
				\toprule
				Comparison with: 
				& \multicolumn{4}{l}{\textbf{Compiler-based Designs}}
				& \multicolumn{2}{l}{\textbf{CNN-to-FPGA Toolflows}}
				& \multicolumn{1}{l}{\textbf{CNN-tailored Designs}} 
				& \multicolumn{1}{l}{\textbf{Overlay Designs}}
				& \multicolumn{1}{l}{\textbf{Cloud-based Designs}}
				& \begin{tabular}[l]{@{}l@{}} \textbf{Interconnect-aware} \\ \textbf{Designs} \end{tabular}
				& \begin{tabular}[l]{@{}l@{}} \textbf{Full-stack-optimised} \\ \textbf{Designs} \end{tabular}
				\\
				
				& Snowflake~\cite{snowflake2017iscas} 
				& \begin{tabular}[l]{@{}l@{}} \tool: \\ ResNet50* \end{tabular}
				& xDNN~\cite{xdnn2020xilinx}
				& DNNVM~\cite{dnnvm2019tcad}
				& \multicolumn{2}{l}{ALAMO~\cite{alamo2020tcad}}
				
				& ResNetAccel~\cite{residaccel2017iscas} 
				& FTDL~\cite{ftdl2020dac}
				& Cloud-DNN~\cite{clouddnn2019fpga}
				& \begin{tabular}[l]{@{}l@{}} Scaling the \\ Cascades~\cite{scaling_the_cascades2019fpl}
				\end{tabular}
				& Full-Stack~\cite{fullstack2021tnnls}
				& \begin{tabular}[l]{@{}l@{}} \tool:\\ ResNet50* \end{tabular}
				\\
				\cmidrule{6-7}
				\midrule
				FPGA  
				& Z7045 
				& Z7045 
				& VU9P 
				& ZU9
				& Arria 10 GX1150
				& Stratix 10 GX2800
				& Arria 10 GX1150 
				& VU125
                & VU9P
                & VU37P
                & Arria 10 GX1150
                & ZU7EV 
				\\
				Clock (MHz) & 250 & 150 & 500 & 500 & 240 & 150 & 300 & 650 & 125 & 650 & 200 & 200 \\
				Precision & 16b fixed & 16b fixed & 8b fixed & 8b fixed & 16b fixed & 16b fixed & 16b fixed & 16b fixed & 16b fixed & 8b fixed & 8b fixed & 16b fixed \\
				DSPs$^\dagger$ & 900 & 900 & 6840 & 2520 & 3036 & 11,520 & 3036 & 1200 & 3036 & 9024 & 3036 & 1728 \\
				Logic Capacity & 218.6 kLUTs & 218.6 kLUTs & 1182.0 kLUTs & 274.0 kLUTs & 427.2 kALMs & 933.0 kALMs & 427.2 kALMs & 716.0 kLUTs & 1182 kLUTs & 1304 kLUTs & 427.2 kALMs & 230.0 kLUTs \\
				Block RAM & 2.40 MB & 2.40 MB & 9.48 MB & 4.01 MB & 6.60 MB & 28.62 MB & 6.60 MB & 11.075 MB & 43.23 MB & 42.61 MB & 6.60 MB & 4.75 MB \\
				{\color{black}DSP Util.$^\dagger$} & 28.4\% & 100\% & 100\% & 83.8\% & 80\% & 80\% & 56.8\% & 100\% & 80.2\% & 95\% & 97\% & 100\% \\
				
				\begin{tabular}[t]{@{}l@{}} 
					inf/s
				\end{tabular} 
				& 17.7
				& 28.18
				& 153.57
				& 80.95
				& 71.38
				& 77.55
				& 33.93
				& 151.22
				& 71.94
				& 766
				& 197.23
				& 71.71
				\\
				\begin{tabular}[t]{@{}l@{}} 
					inf/s/DSP$^\dagger$
				\end{tabular} 
				& 0.0196
				& 0.0313
				& 0.0112
				& 0.016
				& 0.0235
				& 0.0067
				& 0.0111
				& 0.1260
				& 0.0105
				& 0.0424
				& 0.0324
				& 0.0415
				\\
				\begin{tabular}[t]{@{}l@{}} 
					inf/s/Logic
				\end{tabular} 
				& 0.0809
				& 0.1289
				& 0.0649
				& 0.1477 
				& 0.1671
				& 0.0831
				& 0.0794
				& 0.2112
				& 0.0608
				& 0.5874
				& 0.4616
				& 0.3117
				\\

				\bottomrule
				
			\end{tabular} 
			\begin{tablenotes}
				\normalsize
				\item * using OVSF50, ** batch size = 1, $\dagger$ 18$\times$18, 19$\times$18 and 25$\times$18 DSP configurations, inf/s/DSP is adjusted based on precision for fair comparison (multiplied by 0.5 for 8b).
			\end{tablenotes}
			
		\end{threeparttable}
	}
	\vspace{0.2cm}
	\label{tab:resnet50_comparison_table}
\end{table*}

\subsubsection{\textbf{Comparison with Existing FPGA Designs}}
\label{sec:fpga_comparison}

To assess the performance of the proposed framework with respect to existing FPGA work, we perform a number of comparison with a broad range of state-of-the-art works that optimise CNN inference from different aspects. These span accelerators that aggressively apply compiler techniques~\cite{snowflake2017compiling,snowflake2017iscas,xdnn2020xilinx,dnnvm2019tcad}, the highest performing FPGA-based accelerators for sparse~\cite{sparsecnnaccel2019fccm} and lightweight CNNs~\cite{lightopu2020fpga}, a multi-accelerator design that addresses PE underutilisation for SqueezeNet~\cite{maximising2017isca}, a state-of-the-art CNN-to-FPGA toolflow~\cite{alamo2020tcad}, an optimised overlay architecture~\cite{ftdl2020dac}, a highly customised accelerator for residual networks~\cite{residaccel2017iscas}, a cloud-optimised framework~\cite{clouddnn2019fpga}, a CNN accelerator designed in an interconnect-aware manner~\cite{scaling_the_cascades2019fpl} and an accelerator that applies full-stack optimisations~\cite{fullstack2021tnnls}. 

Table~\ref{tab:comparison_table} lists the performance results for ResNet18/34 and SqueezeNet. On Z7045, \tool achieves 2.33$\times$ and 1.12$\times$ higher throughput than \cite{snowflake2017compiling} and \cite{sparsecnnaccel2019fccm}, respectively. For SqueezeNet, our design delivers 1.83$\times$ and 1.67$\times$ higher performance density in inf/s/DSP and inf/s/Logic than Light-OPU~\cite{lightopu2020fpga}. 
Compared to the multi-accelerator design~\cite{maximising2017isca} that also addresses the PE underutilisation, \tool yields 1.40$\times$ higher inf/s/DSP and 1.14$\times$-1.27$\times$ higher inf/s/Logic despite having the same (V48T-based design~\cite{maximising2017isca}) or 36\% lower (V690T-based design~\cite{maximising2017isca}) on-chip memory budget.

The original ResNet50 reaches 76.15\% accuracy with a model size of 25.56M parameters. With \tool's ResNet50-OVSF50 variant improves accuracy to 76.23\% while having 10\% fewer parameters (22.84M).
Table~\ref{tab:resnet50_comparison_table} presents the measured performance results for ResNet50. On Z7045, \tool outperforms Snowflake by 1.59$\times$ in inf/s. Compared with designs on larger devices, our design achieves higher performance density (inf/s/DSP) by 3.69$\times$, 2.58$\times$, 1.76$\times$-6.16$\times$, 3.17$\times$ and 3.94$\times$ over xDNN, DNNVM, ALAMO, ResNetAccel and Cloud-DNN. The overlay-based FTDL reaches higher inf/s/DSP and 1.47$\times$ lower inf/s/Logic, but targets a platform with 2.33$\times$ larger on-chip memory and 2$\times$ higher bandwidth, both of which substantially reduce the off-chip memory accesses and the associated latency. Similarly, compared to the interconnect-aware design of \cite{scaling_the_cascades2019fpl}, \tool reaches 97.8\% of its inf/s/DSP, despite using a platform with 8.9$\times$ smaller on-chip memory. Finally, \tool outperforms the full-stack-optimised accelerator of \cite{fullstack2021tnnls} by 1.28$\times$ in inf/s/DSP.

\textbf{Discussion.}
Based on the presented evaluation, \tool consistently outperforms a wide range of FPGA-based accelerator designs, in spite of their diverse designs.
As such, our framework delivers an average throughput gain of 2.23$\times$ (2.05$\times$ geo. mean) over designs that aggressively apply compiler optimisations on fixed accelerators~\cite{snowflake2017compiling,snowflake2017iscas,xdnn2020xilinx,dnnvm2019tcad} and, at the same time, achieves an average inf/s/DSP gain of 2.5$\times$ (2.41$\times$ geo. mean) over highly customised CNN-tailored designs~\cite{lightopu2020fpga, residaccel2017iscas} and 3.94$\times$ over the cloud-optimised mapping of Cloud-DNN.
A notable comparison is with the sparse CNN accelerator for ResNet34 presented in~\cite{sparsecnnaccel2019fccm}, with \tool achieving 12\% throughput gain. It should be noted that the sparse CNN accelerator applies Deep Compression~\cite{deepcompression2015iclr} to sparsify the target CNN, employs a specialised dataflow and modifies the underlying PEs in order to extract high performance. 
In contrast, \tool improves the performance of CNN engines while affecting neither the selected dataflow nor the internal design of the PEs, and still delivers 12\% higher throughput than the sparse CNN accelerator. 

\begin{table}[b]
\rev{
	\centering
    \vspace{0.3cm}
    \captionsetup{font=small,labelfont=bf}
	\caption{\rev{\footnotesize Resource usage breakdown of \tool's designs.}}
    \vspace{-0.2cm}
	\resizebox{0.65\linewidth}{!}{%
		\begin{tabular}{l l l l l l}
			\toprule
			\multicolumn{1}{l}{\multirow{1}{*}{\textbf{Design Config.}}} & \textbf{Platform} & \textbf{Resource Type} & \multicolumn{1}{c}{\multirow{1}{*}{\textbf{\texttt{CNN-WGen}}}} & \textbf{CNN Engine} \\
			\midrule
			\multirow{2}{*}{ResNet18-OVSF50} & \multirow{2}{*}{ZC706} 
			& DSPs & 7.5\% & 92.5\% \\
            & & LUTs & 1\% & 74\% \\
	          \midrule		
            \multirow{2}{*}{ResNet34-OVSF50} & \multirow{2}{*}{ZC706}  
			& DSPs & 11.3\% & 88.7\% \\
            & & LUTs & 3\% & 75\% \\
            \midrule
            \multirow{2}{*}{ResNet50-OVSF50} & \multirow{2}{*}{ZC706}  
			& DSPs & 11.1\% & 88.9\% \\
            & & LUTs & 3\% & 75\% \\
			\bottomrule
		\end{tabular}%
	}
	\label{tab:rsc_usage}
    }
\end{table}

\subsubsection{\textbf{Resource Usage}}
\label{sec:rsc_usage}
We select \tool and baseline designs with up to 1-pp accuracy drop and compare their post place-and-route resource usage on Z7045, reported in [DSPs, BRAM, LUTs] tuples for 4$\times$ bandwidth.
For ResNet34, the faithful baseline consumes {\small [99\%,83\%,77\%]}, Tay82 {\small [99\%,79\%,77\%]}, OVSF50 {\small [100\%,81\%,78\%]} and Tay82+OVSF50 {\small [100\%,87\%,81\%]}. For ResNet18, the faithful baseline {\small [78\%,99\%,70\%]}, Tay88 {\small [78\%,99\%,69\%]}, OVSF50 {\small [100\%,87\%,75\%]} and Tay82+OVSF50 {\small [100\%,83\%,80\%]}. For ResNet50, OVSF50 on ZU7EV consumes {\small [100\%,87\%,78\%]}. Finally, the input selective PE mechanism adds a minimal LUTs overhead of less than 7\%. \rev{We further report the breakdown of resource consumption between \texttt{CNN-WGen} and the CNN Engine in Table~\ref{tab:rsc_usage}. We observe that, using our performance model, the DSE stage is able to balance the allocation of DSPs between the two modules. Moreover, the LUT overhead of the weights generator is minimal compared to the CNN Engine, providing a beneficial trade-off.}

\begin{figure}[t]
    \centering
    \begin{subfigure}{0.45\textwidth}
        \centering
        \includegraphics[width=1.25\columnwidth,trim={10cm 4.5cm 6cm 6.2cm},clip]{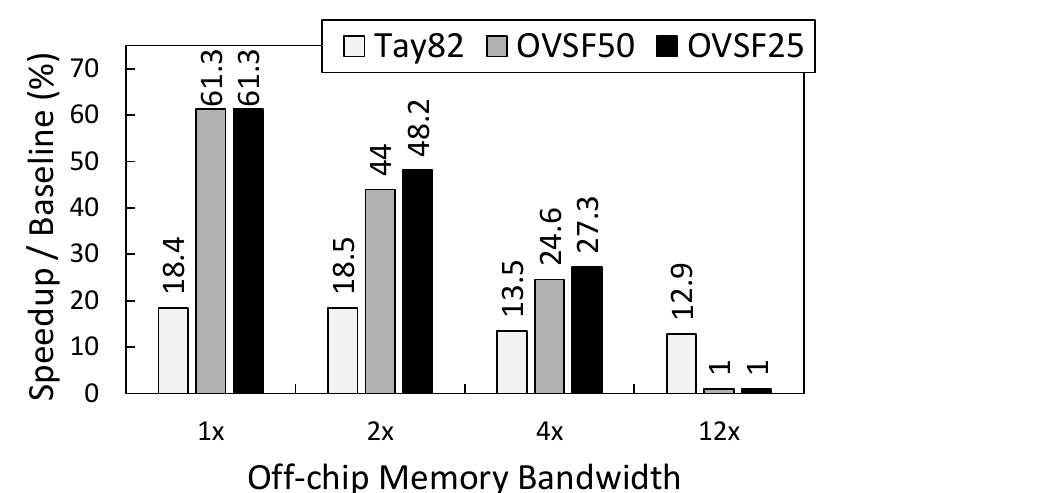}
        \caption{ResNet18 - Z7045}
        \label{fig:resnet18z7045}
    \end{subfigure}
    \begin{subfigure}{0.45\textwidth}
        \centering
        \includegraphics[width=1.25\columnwidth,trim={10cm 4.5cm 6cm 6.2cm},clip]{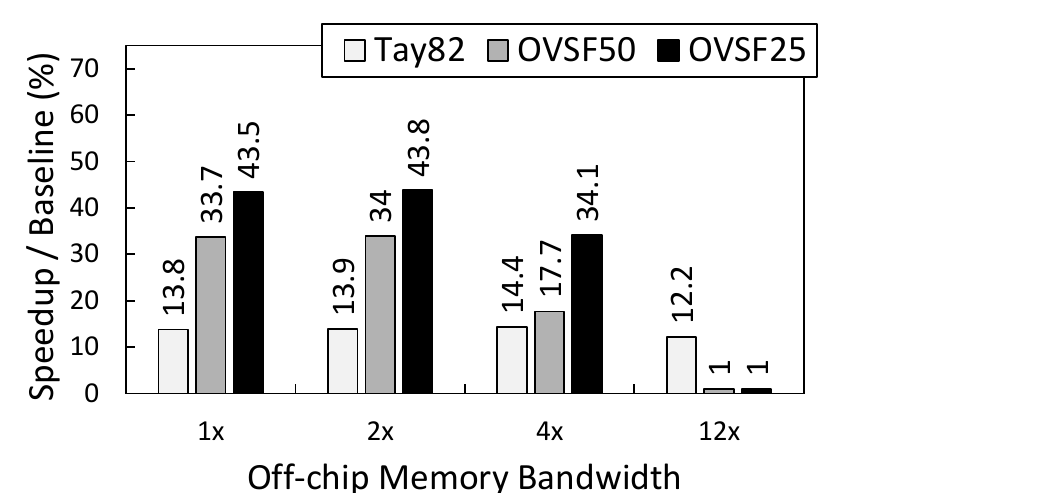}
        \caption{ResNet18 - ZU7EV}
        \label{fig:resnet18zu7ev}
    \end{subfigure}
    \begin{subfigure}{0.45\textwidth}
        \centering
        \includegraphics[width=1.25\columnwidth,trim={10cm 4.5cm 6cm 6.2cm},clip]{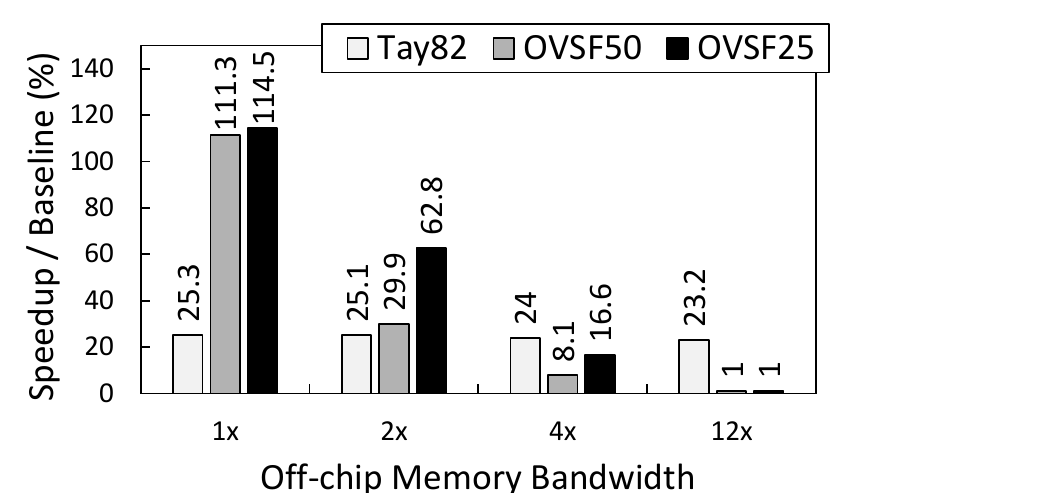}
        \caption{ResNet34 - Z7045}
        \label{fig:resnet34z7045}
    \end{subfigure}
    \begin{subfigure}{0.45\textwidth}
        \centering
        \includegraphics[width=1.25\columnwidth,trim={10cm 4.5cm 6cm 6.2cm},clip]{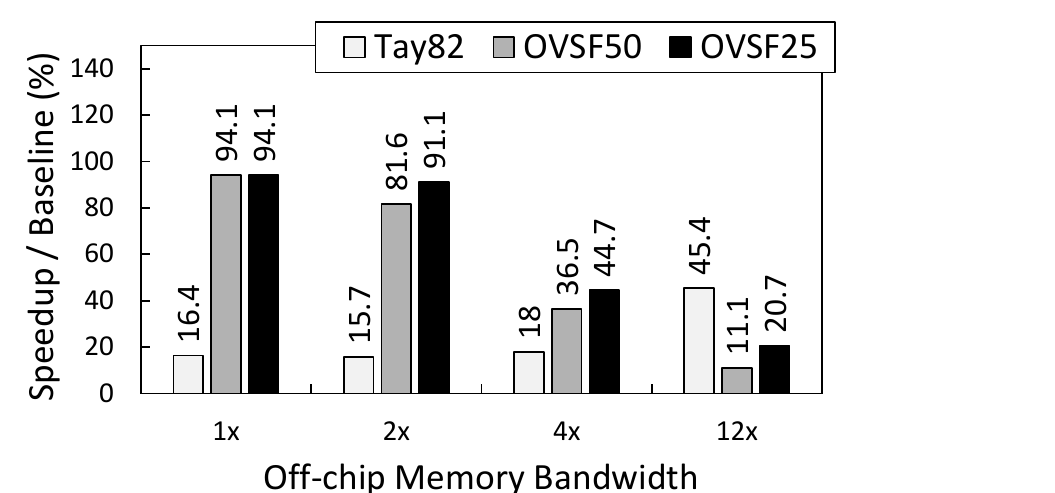}
        \caption{ResNet34 - ZU7EV}
        \label{fig:resnet34zu7ev}
    \end{subfigure}
    \vspace{-0.2cm}
    \captionsetup{font=small,labelfont=bf}
    \caption{\footnotesize Speedup over optimised baselines when varying the available off-chip memory bandwidth. 
    }
    \label{fig:bw_sensitivity}
\end{figure}

\subsection{Sensitivity to Off-Chip Memory Bandwidth}
\label{sec:mem_bandwidth_eval}
Fig.~\ref{fig:bw_sensitivity} shows the impact of varying off-chip memory bandwidth over performance on the two target platforms. The figure compares the speedup of \tool and the Tay82 baseline over the vanilla baseline when varying the external memory bandwidth from 1$\times$ to 12$\times$. The bandwidth's impact is most prominent on the larger ZU7EV, where the performance gains are sustained higher across 1$\times$-4$\times$. In the case of the mid-tier Z7045, we observe a sharper drop in the speedup as the bandwidth increases. This is due to the more limited computational resources of Z7045, which makes most CNN layers compute-bound. In contrast, the abundance of computational resources on ZU7EV makes the CNN layers more memory-bound. For instance, at 4$\times$ bandwidth (4.5 GB/s), the vanilla ResNet18 baseline yields DSP utilisation of 71\% on the compute-bound Z7045 and 53\% on the more memory-bound ZU7EV. In this case, \tool significantly improves both cases by mapping ResNet18-OVSF25 with 89\% and 71\% DSP utilisation.
As a result, \tool sustains its gains across a wider range of bandwidths and outperforms Tay82, until the bandwidth-abundant case (12$\times$) where computational resources become the critical factor. In this case, Tay82's lower number of operations due to pruning leads to higher performance.

Across the designs, the input selective PEs contribute an additional speedup of up to 20\%, with varying gains depending on the CNN-FPGA pair and the available bandwidth. For ResNet34-OVSF25 on ZU7EV, disabling this mechanism leads to 0/13.9/3.3/5.9\% lower throughput for the four bandwidths, with a similar pattern observed for the rest of the CNNs. Our input selective PEs effectively improve the performance of suboptimally mapped layers in compute-bound settings, whereas no gain is obtained for the most bandwidth-constrained case (1$\times$) where the designs are severely memory-bound, limiting further improvements through higher PE utilisation.

\subsection{Impact of Input Selective PEs}
\label{sec:input_sel_pes_eval}
Here, we evaluate the impact of input selective PEs on the achieved performance.
This is investigated by implementing \tool's selected hardware design for each of the benchmark CNNs with and without the input selective PEs and comparing the achieved performance, measured on the two target FPGA platforms. When the input selective PEs are omitted, we call the designs \textit{ablated}. 
Table~\ref{tab:input_sel_pes_eval} presents the achieved performance gains between the two designs.

\begin{table}[t]
    \small
    \centering
    \captionsetup{font=small,labelfont=bf}
    \caption{Ablation study of input selective PEs.}
    \vspace{-0.1cm}
    \resizebox{0.5\linewidth}{!}{
    \begin{tabular}{l l l r l c}
        \toprule
        \multicolumn{1}{l}{Model} & & FPGA & \multicolumn{2}{c}{Input Selective PEs} & Performance \\
        \multicolumn{1}{l}{Arch.} & & Platform & \multicolumn{1}{c|}{without} & with & Gain \\
        \midrule
        
        & OVSF50 & Z7045 & 49.9 inf/s & 49.9 inf/s & 1.00$\times$ \\
        \multicolumn{1}{l}{\multirow{1}{*}{ResNet18}} & OVSF25 & Z7045 & 50.4 inf/s & 51.0 inf/s & 1.01$\times$ \\
        \cmidrule{2-6}
        & OVSF50 & ZU7EV & 124.1 inf/s & 124.1 inf/s & 1.00$\times$ \\
        & OVSF25 & ZU7EV & 135.2 inf/s & 135.2 inf/s & 1.00$\times$ \\
        
        \midrule
        
        & OVSF50 & Z7045 & 25.4 inf/s & 31.1 inf/s & 1.22$\times$ \\
        \multicolumn{1}{l}{\multirow{1}{*}{ResNet34}} & OVSF25 & Z7045 & 33.5 inf/s & 33.3 inf/s & \phantom{1}0.6\% \\
        \cmidrule{2-6}
        & OVSF50 & ZU7EV & 81.1 inf/s & 81.1 inf/s & 1.00$\times$ \\
        & OVSF25 & ZU7EV & 72.4 inf/s & 88.0 inf/s & 1.21$\times$ \\
        
        \midrule
        
        & OVSF50 & Z7045 & 23.7 inf/s & 27.0 inf/s & 1.14$\times$ \\
        \multicolumn{1}{l}{\multirow{1}{*}{ResNet50}} & OVSF25 & Z7045 & 23.7 inf/s & 28.1 inf/s & 1.18$\times$ \\
        \cmidrule{2-6}
        & OVSF50 & ZU7EV & 63.1 inf/s & 71.7 inf/s & 1.13$\times$ \\
        & OVSF25 & ZU7EV & 68.5 inf/s & 77.8 inf/s & 1.13$\times$ \\
        
        \midrule
        
        \multicolumn{1}{l}{\multirow{1}{*}{SqueezeNet}} & OVSF50 & ZU7EV & 724.2 inf/s & 792.2 inf/s & 1.09$\times$ \\
        & OVSF25 & ZU7EV & 731.4 inf/s & 800.6 inf/s & 1.09$\times$ \\
        \midrule
        \midrule
        Average & & & & & 1.12$\times$ \\
        Geo. Mean & & & & & 1.11$\times$ \\
        \bottomrule
    \end{tabular}
    }
    \vspace{0.2cm}
    \label{tab:input_sel_pes_eval}
\end{table}

The PE-enhancing mechanism contributes varying throughput gains, yielding up to 22\% faster inference and an average improvement of 12\% (11\% geo. mean). 
For ResNet18, the ablated designs already sustain high DSP utilisation with ResNet18-OVSF50 and -OVSF25, reaching 90\% and 86.5\% of the theoretical peak performance of Z7045 and ZU7EV, respectively.
On the other hand, the ablated ResNet34-OVSF50 design on Z7045 achieves only 69.6\% of the theoretical peak throughput. Similarly, the ablated ResNet34-OVSF25 design on ZU7EV achieves 77.5\% of the theoretical performance. In both cases, the input selective PEs are able to substantially increase the DSP utilisation, with the enhanced CNN engines achieving 85.1\% and 94.2\% of the peak performance, respectively.

A similar effect is observed for ResNet50 and SqueezeNet. The ablated designs yield 73.8\% of the theoretical peak performance for both OVSF50 and OVSF25 on Z7045, and 76.7\% and 83.3\% for OVSF50 and OVSF25, respectively, on ZU7EV. In this case, our input selective PEs are able to improve the DSP utilisation, achieving 84.1\% and 87.4\% of the peak throughput on Z7045 for OVSF50 and OVSF25, respectively, and 87.2\% and 94.7\% for OVSF50 and OVSF25, respectively, on ZU7EV. Finally, for SqueezeNet, the input selective PEs improve the measured throughput from 73.2\% to 80.1\% of the peak performance for OVSF50 and from 73.9\% to 80.9\% for OVSF25 on ZU7EV. As such, enhancing a CNN engine's PEs with our proposed input selectivity technique alleviates the resource underutilisation due to the diverse layer shapes within a CNN. In the cases where our technique is estimated to provide minimal gains (\textit{i.e.} $<$5\%) and its usage is not justified, \tool opts for omitting it to save LUT resources.


\begin{figure}[t]
    \centering
    \begin{subfigure}{0.45\textwidth}
        \centering
        \includegraphics[width=1.0\columnwidth,trim={0cm 0cm 0cm 0cm},clip]{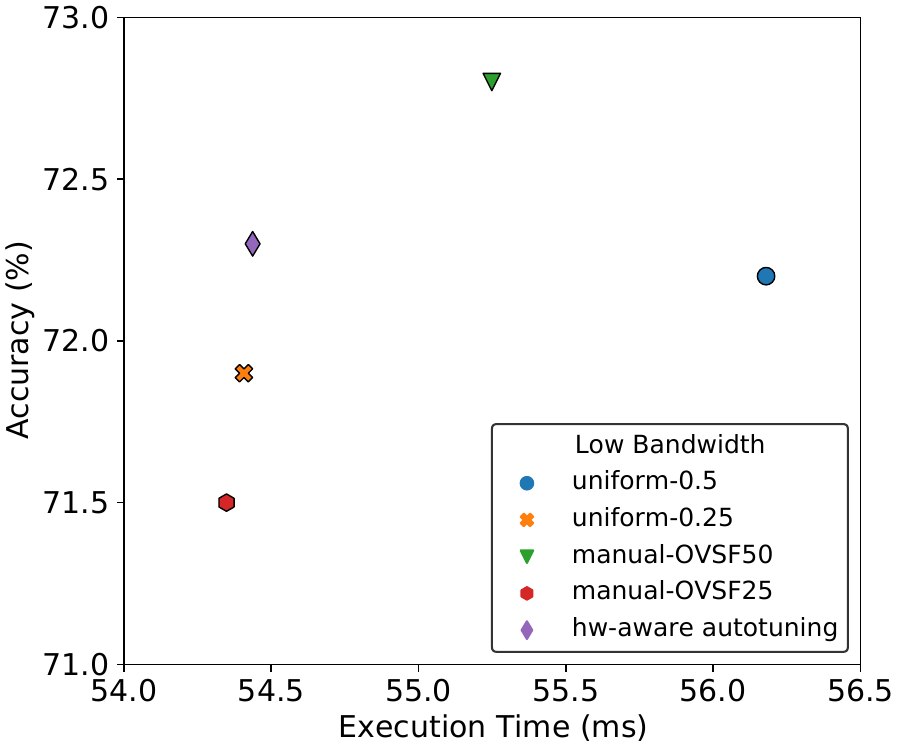}
        \caption{ResNet34 - Z7045 with Low Bandwidth}
        \label{fig:hw_autotune_resnet34_low_bw}
    \end{subfigure}
    \begin{subfigure}{0.45\textwidth}
        \centering
        \includegraphics[width=1.0\columnwidth,trim={0cm 0cm 0cm 0cm},clip]{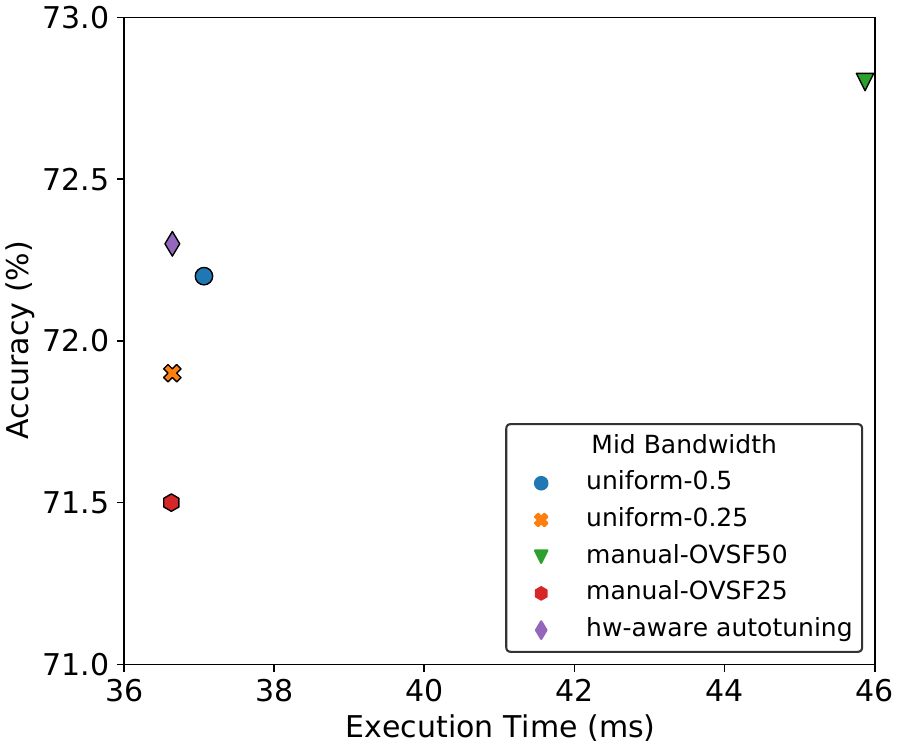}
        \caption{ResNet34 - Z7045 with Medium Bandwidth}
        \label{fig:resnet18zu7ev}
    \end{subfigure}
    \hfill
    \begin{subfigure}{0.45\textwidth}
        \centering
        \includegraphics[width=1.0\columnwidth,trim={0cm 0cm 0cm 0cm},clip]{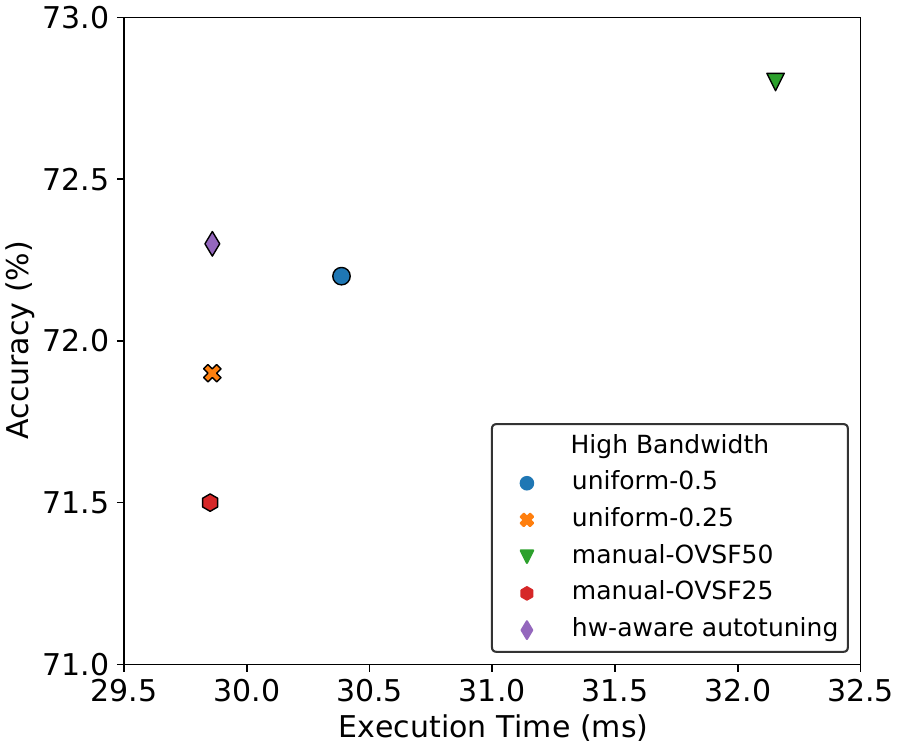}
        \caption{ResNet34 - Z7045 with High Bandwidth}
        \label{fig:resnet34z7045}
    \end{subfigure}
    \vspace{-0.2cm}
    \captionsetup{font=small,labelfont=bf}
    \caption{\footnotesize Accuracy-execution time trade-off for different OVSF ratio selections. Hardware-aware autotuning closes the performance gap between OVSF50 and OVSF25 while being within 1pp of the original model's accuracy (73.3\%). 
    }
    \label{fig:bw_sensitivity}
\end{figure}

\subsection{Hardware-Aware vs. Manual Tuning of OVSF Ratios}
\label{sec:eval_hw_aware_ratios}

Next, we evaluate the effectiveness of our hardware-aware tuning of OVSF ratios in yielding designs with improved accuracy-performance trade-off. To this end, we compare against two ratio selection methods: \textit{i)}~\texttt{uniform-$\rho$} which uses the same ratio $\rho$ across all layers, with the exception of the first CONV layer. This baseline represents a brute-force approach of setting the OVSF ratios; and \textit{ii)}~\texttt{manual-OVSF50} and \texttt{manual-OVSF25} which use the manually selected ratios detailed in Section~\ref{sec:training_scheme} to achieve 50\% and 75\% reduction in model size from the original model. This baseline constitutes an optimised hand-engineered method. We perform the comparison by implementing ResNet18 and ResNet34 using both the hardware-aware and the baseline flows for different bandwidth availability and comparing the achieved performance, measured on Z7045.

Figure~\ref{fig:hw_autotune_resnet34_low_bw} shows the achieved accuracy and execution time measured on the target FPGA and depicts how our method, denoted by \texttt{hw-aware autotuning}, yields Pareto-optimal designs that were previously unattainable. For ResNet34, our method sustains the same performance as the fast OVSF25 design across all memory bandwidths. However, it additionally improves OVSF25's accuracy by 0.8pp, thus outputting design that are within 1pp of the original model's accuracy (72.3\% for all three bandwidths vs 73.3\% for the vanilla ResNet34). At the same time, it is consistently faster than the coarse \texttt{uniform-0.5}. 
We obtain similar results for ResNet18, with the same processing speed as OVSF25 and accuracy gains in the range of 0.3pp-1.2pp (0.86pp average gain across bandwidths) over OVSF25. Across all cases, the \texttt{uniform-$\rho$} baselines were either unnecessarily slow (\texttt{uniform-0.5}) or low in accuracy (\texttt{uniform-0.25}), further advocating for a principled method of selecting the OVSF ratios.

By exploiting the bounding factor of each layer, our hardware-aware scheme selectively allows for a longer weights generation stage without affecting the processing speed. As such, we can obtain a better approximation of the weights and sustain high throughput. As shown through our experiments, the hardware-aware methodology yielded competitive designs, performing either better or in par even against highly optimised hand-tuned configurations (OVSF50 and OVSF25).
\stelios{Added this.}

\subsection{Comparison with Embedded GPU}
\label{sec:gpu}

With the majority of CNNs deployed for inference on embedded and mobile devices, our evaluation focuses on the embedded space. In power-constrained applications, the main metrics of interest comprise: \textit{1)}~the absolute power consumption and \textit{2)}~the energy efficiency in performance-per-watt. In this respect, we investigate the energy efficiency of \tool in relation to the widely used high-performance NVIDIA Jetson TX2 platform. In all cases, for \tool we use the OVSF50 variant with less than 1-pp accuracy drop.

For the performance evaluation on TX2, we use NVIDIA TensorRT as supplied by the JetPack~4.1.1 package. TensorRT is run with the NVIDIA cuDNN library and 16-bit half-precision floating-point arithmetic (FP16), which enables the highly optimised execution of layers. Across all platforms, each CNN is run 100 times to obtain the average throughput. Furthermore, power measurements for the GPU and FPGAs are obtained via a power monitor on the corresponding board. In all cases, we subtract the average idle power from the measurement to obtain the power due to benchmark execution. The idle power of the FPGA platforms is measured at the board level with no design programmed in the FPGA fabric, so that the clock tree power and the power leakage of the chip are also included in the run-time power due to benchmark execution.
Across all experiments, we used a batch size of 1, as is typical in mobile and embedded settings.

\begin{figure}[t]
    \centering
    \vspace{-0.4cm}
    {
    \includegraphics[width=0.65\textwidth,trim={2cm 8cm 1cm 8.5cm},clip]{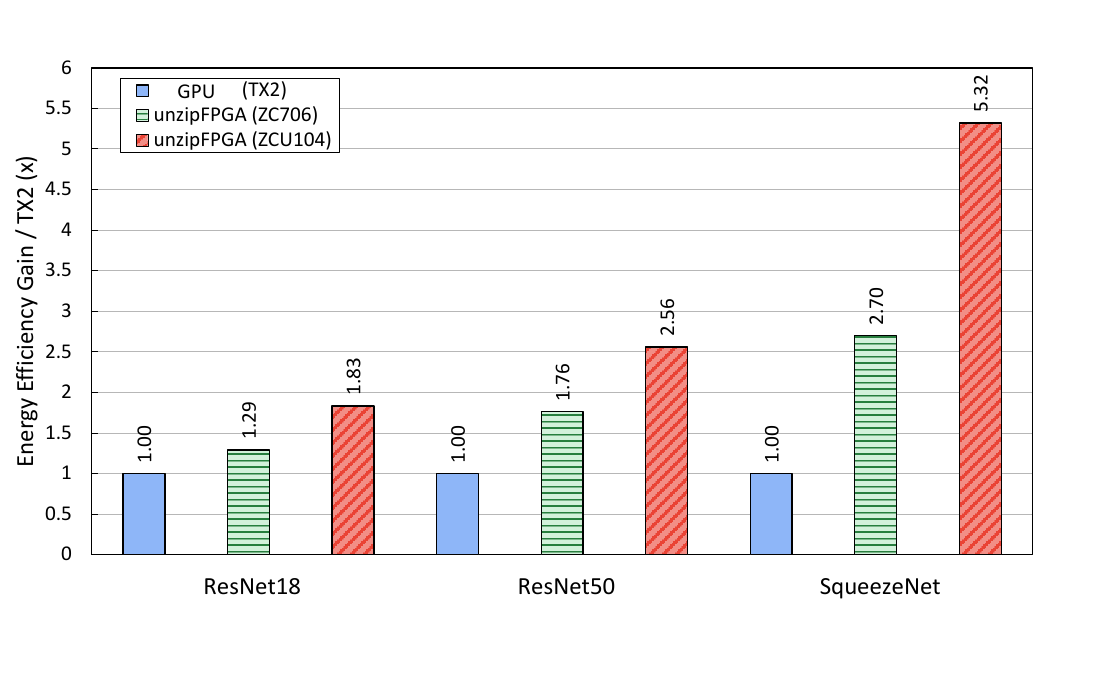}
    }
    \vspace{-0.8cm}
    \captionsetup{font=small,labelfont=bf}
    \caption{\footnotesize Energy efficiency comparison between \tool and TX2 designs.}
    \vspace{0.2cm}
    \label{fig:gpu_comparison}
\end{figure}

Tegra X2 mounts a 256-core GPU with native support for FP16 arithmetic which supports a range of operating modes with different clock frequencies and power consumption. To perform a fair comparison with respect to energy efficiency in terms of performance-per-watt, we configure the GPU with the maximum energy efficiency mode (Max-Q) which sets the frequency of the GPU at 850~MHz and configures all components of TX2 to achieve the best power-throughput trade-off. Fig.~\ref{fig:gpu_comparison} presents the conducted comparison. \tool achieves an energy efficiency improvement over TX2 of up to 5.32$\times$ in inf/s/W with an average of 2.57$\times$ (2.31$\times$ geo. mean) across the benchmarks. As a result, \tool consistently demonstrates significant gains in performance-per-watt across the benchmarks over highly optimised embedded GPU implementations.

\section{Conclusion}
\label{sec:conclusion}

In this work we have presented \tool, a framework for FPGA-based CNN accelerators that mitigates the limitations that prevent single computation engines from attaining high resource utilisation and throughput. By generating the layer weights on demand and selectively balancing the PE load, \tool outperforms both status-quo and pruned CNN engines for the same bandwidth, while largely improving performance density compared to diverse state-of-the-art CNN accelerators. Furthermore, we demonstrated the superiority of models optimised with \tool in terms of energy efficiency compared to these being deployed on embedded GPU platforms. 
The benefits of the proposed on-the-fly formulation brought the largest gains at reduced memory bandwidths, which we envision to be a turning point towards enabling multi-tenant FPGA-based CNN models running concurrently and sharing the same off-chip memory. 

\bibliographystyle{ACM-Reference-Format}
\bibliography{references}

\end{document}